%% file: main.tex
\documentclass[10pt,twocolumn,letterpaper]{article}

\usepackage{cvpr}              
\input{preamble}
\definecolor{cvprblue}{rgb}{0.21,0.49,0.74}
\usepackage[pagebackref,breaklinks,colorlinks,allcolors=cvprblue]{hyperref}

\def\paperID{6} 
\def\confName{CVPR}
\def\confYear{2026}

\title{Benchmarking Single-Step Inpainting Methods for Multi-Object 3D Gaussian Splatting Scenes}

\author{
    Finn Dröge\textsuperscript{1}\quad
    Cecilia Curreli\textsuperscript{1,2}\quad
    Abhishek Saroha\textsuperscript{1,2}\quad
    Daniel Cremers\textsuperscript{1,2} \\
    \\
    \textsuperscript{1}Technical University of Munich \quad
    \textsuperscript{2}Munich Center for Machine Learning \\
}

\usepackage{xspace}
\usepackage{dsfont}

\newcommand{\InitNanoBanana}{Init-NanoBanana\xspace}
\newcommand{\InitNanoBananaShort}{Init-NB}

\newcommand{\Init}{Init-2DInpainter\xspace}
\newcommand{\InitLaMa}{Init-LaMa\xspace}
\newcommand{\InitLaMaShort}{Init-LM}
\newcommand{\InitPowerPaint}{Init-PowerPaint\xspace}
\newcommand{\InitPowerPaintShort}{Init-PP}

\newcommand{\FTThree}{Finetune-2DInpainter~(3)\xspace}
\newcommand{\FTLaMaThree}{Finetune-LaMa~(3)\xspace}
\newcommand{\FTLaMaThreeShort}{FT-LM~(3)}

\newcommand{\FTPowerPaintThreeShort}{FT-PP~(3)}

\newcommand{\FTAll}{Finetune-2DInpainter~(All)\xspace}
\newcommand{\FTLaMaAll}{Finetune-LaMa~(All)\xspace}
\newcommand{\FTLaMaAllShort}{FT-LM~(All)}

\newcommand{\FTPowerPaintAllShort}{FT-PP~(All)}

\begin{document}
\maketitle
\input{sec/0_abstract}    
\input{sec/1_intro}

\input{sec/3_method}

\input{sec/4_experiments}
\input{sec/5_conclusion}

{
    \small
    \bibliographystyle{ieeenat_fullname}
    \bibliography{main}
}

\input{sec/X_suppl}

\end{document}

%% file: sec/0_abstract.tex
\begin{abstract}
The tasks of object removal and inpainting 3D Gaussian Splatting (3DGS) scenes face challenges such as 3D consistency across camera views.
In comparing 2D inpainters and their suitability for the 3D domain, we find that reconstruction-based inpainters outperform generative diffusion models in 3D consistency. Integrating these 2D inpainters into different single-step methods for creating and finetuning 3DGS scenes, our results indicate that initializing the scene from scratch produces higher quality results than finetuning the existing scene. Using a state-of-the-art generative 2D inpainter, we create a straightforward baseline to underline the importance of object removal before inpainting in the 3D setting.
Since 360° datasets rarely include real-world ground truths, and challenging occlusion scenarios are equally sparse, we introduce a novel multi-object scene with recorded ground truth data and many views with object occlusions.
\end{abstract}

%% file: sec/1_intro.tex
\section{Introduction}
\label{sec:intro}

Neural Radiance Fields (NeRFs) and 3D Gaussian Splatting (3DGS) have revolutionized 3D reconstruction of scenes. While the reconstruction of static scenes is very accurate in vanilla versions of the methods~\cite{vanilla_3DGS, nerf}, applications involving scene editing and animation still have issues~\cite{3dgic, gaussianstolife}. In particular, a simple object removal results in a hole in the scene. In 2D, this issue can be resolved with an inpainting model. To accurately model the world, 3D inpainters have to not only focus on visual fidelity but also ensure consistency across all views, posing a more difficult problem.

While methods like 3DGIC~\cite{3dgic} iteratively refine the 2D inpainting results of the same view with LDM~\cite{ldm} during scene optimization, we choose to investigate a more straightforward line of work, which inpaints a given image only once. We call these methods single-step.
In this work, we conduct a comprehensive benchmark to assess how effectively various single-step inpainting methods can translate 2D inpainting results into 3D. We introduce three methods built on top of Gaussian Grouping~\cite{gaussian_grouping} and 3DGIC~\cite{3dgic}.

Motivated by recent advances in 2D inpainting methods~\cite{LaMa, powerpaint, brushnet, nanobanana}, we investigate whether the quality of images generated with these state-of-the-art inpainting models is sufficient to redefine the 3DGS inpainting paradigm~\cite{painpainter, nerfiller}. The current inpainting pipeline~\cite{3dgic, gaussian_grouping} first removes the object from the scene and then refines it by optimizing over a set of inpainted images, obtained by a 2D inpainter and an inpainting mask over the created hole. It seems natural to ask whether the inpainted 3DGS scene can be obtained by inpainting the input images \textit{directly} in 2D and initializing the scene based on these inpainted results. We investigate this straightforward approach with the state-of-the-art 2D inpainter Nano Banana~\cite{nanobanana}.

The field of 3D inpainting is not only focused on more performant and consistent methods, but also on defining robust evaluation pipelines. While existing datasets like the SPIn-NeRF dataset~\cite{spinnerf} have a real-world ground truth, they mainly consist of forward-facing scenes. Conversely, datasets like Mip-NeRF 360~\cite{mipnerf360} and InNeRF360~\cite{innerf360} demonstrate scenes from all angles, but lack a real-world ground truth for the object removal task. While these datasets~\cite{mipnerf360, innerf360} provide a synthetic ground truth for each view, a recorded real-world ground truth represents reality more accurately. We propose a new scene with a challenging occlusion scenario and benchmark our methods on this scene and two widely employed ones. \\ 
Our main contributions are summarized as follows:
\begin{itemize}[topsep=0px]
    \item \textbf{Comparative Analysis of Single-Step Methods}. We compare various single-step 3D inpainting pipelines and 2D inpainting models and evaluate their suitability for the 3D setting. Generative models~\cite{powerpaint, nanobanana} produce higher fidelity inpaintings in 2D but struggle more with aligning the views in the 3D setting in comparison to LaMa~\cite{LaMa}.
    \item \textbf{Novel Multi-Object Occlusion Scene "Living Room"}. To improve multi-object scene evaluation, we introduce a new 360° living room scene with ground truth data. It includes multiple objects in front of and behind the to-be-removed object, giving a challenging occlusion scenario.
    \item \textbf{Diagnostic Baseline for Pipeline Architecture}. We present a straightforward baseline method that inpaints with Nano Banana~\cite{nanobanana} to demonstrate the importance of object removal before inpainting in the pipeline. By inpainting on the object directly, the method struggles particularly with the occlusion scenario of our novel scene.
\end{itemize}

%% file: sec/3_method.tex
\section{Method}
\label{sec:method}
We present the general 3D inpainting pipeline and a notation. The final scene can be created through finetuning the existing scene or initializing the scene from scratch.

\subsection{3D Scene Inpainting Pipeline}
\label{subsec:preliminaries}
\begin{figure*}[t]
    \centering
    \includegraphics[width=1\linewidth]{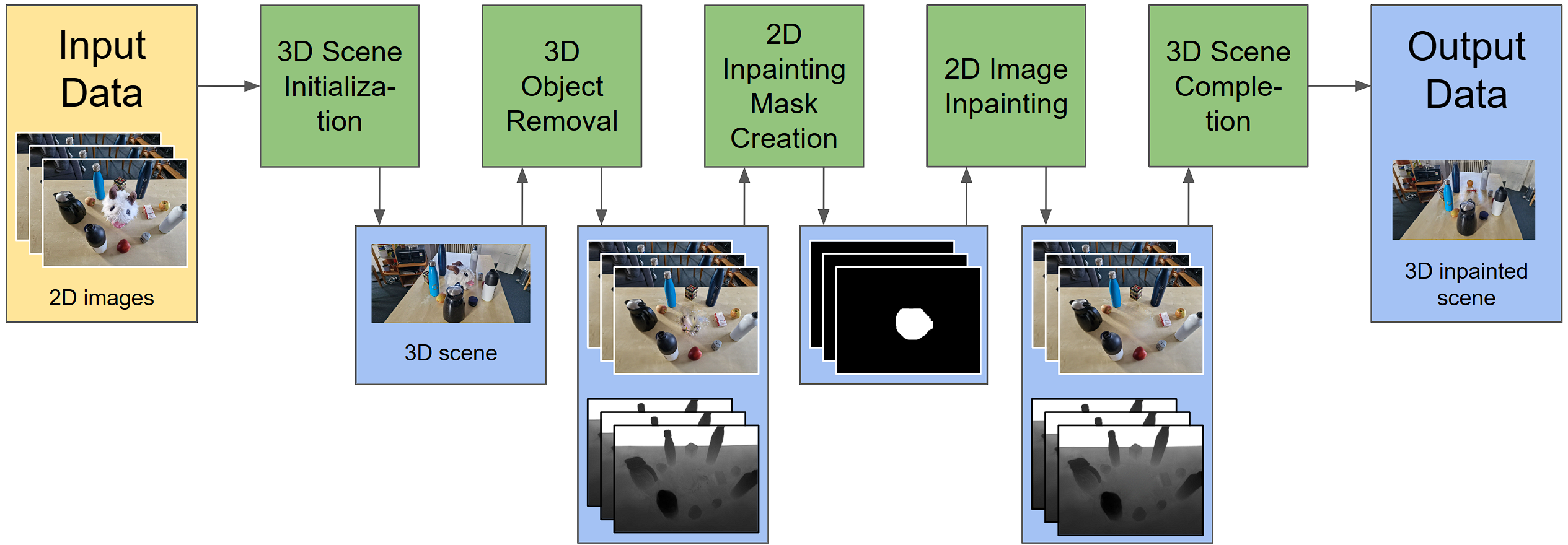}
    \caption{\textbf{Overview of the 3D Scene Inpainting Pipeline}. First, a 3D Gaussian Splatting scene is initialized from input images using Gaussian Grouping~\cite{gaussian_grouping}. Then, the object is removed using the identity encoding of the Gaussians. Using the object masks, the inpainting masks are created for the resulting hole in the scene using a method from 3DGIC~\cite{3dgic}. Using a 2D inpainter the images and depth images are inpainted and finally used for finetuning the scene and covering the hole.}
    \label{fig:model_overview}
\end{figure*}

The general pipeline can be divided into five sub-tasks as shown in the green boxes in \cref{fig:model_overview}.

\textbf{3D Scene Initialization}. First, we use Gaussian Grouping~\cite{gaussian_grouping} to create a 3DGS scene with an additional semantic parameter for each Gaussian.

\textbf{3D Object Removal}. With this semantic parameter, we create object masks and remove all Gaussians of the object.

\textbf{2D Inpainting Mask Creation}. We use the method introduced by 3DGIC~\cite{3dgic} to create the inpainting masks for the hole in the scene rather than the entire object.

\textbf{2D Image Inpainting}. Using these inpainting masks, we inpaint the images and depth images from the object removal step with a~2D inpainter~\cite{LaMa, powerpaint, brushnet}.

\textbf{3D Scene Completion}. There are two different variants to completing the scene. The first option is to finetune the existing scene and fill the hole in 3D. Alternatively, we present a novel method, which takes the inpainted images and initializes the entire scene from scratch.

\textbf{Notation}. We denote methods as \textit{[Finetune $\mid$ Init]-2DInpainter~(NumberOfReferenceViews)}, where \textit{Finetune} (\textit{FT}) describes finetuning in the final step, \textit{Init} initializes the scene from scratch,~\textit{2DInpainter} specifies the 2D inpainter used, and when finetuning we state the number of reference views (\textit{3} or \textit{All}). As~2D inpainters, we use \textit{LaMa}~(\textit{LM})~\cite{LaMa}, \textit{NanoBanana}~(\textit{NB})~\cite{nanobanana}, or \textit{PowerPaint}~(\textit{PP})~\cite{powerpaint}.

\subsection{Finetune the Existing Scene}
Similar to other works~\cite{3dgic, gaussian_grouping}, we finetune on the existing scene without the object and add additional Gaussians to fill the hole. We simplify their~\cite{3dgic, gaussian_grouping} losses by optimizing the masked region with the 2D inpainted results rather than introducing projection-based cross-view consistent losses. Reference views have an RGB~$\mathcal{L}_{1}$ loss and non-reference views a perceptual LPIPS loss~\cite{LPIPS}. Since the background already has a high quality before finetuning, we freeze the weights of the Gaussians from the original scene and only optimize over the new ones.
We introduce two different finetuning methods with 3 and all views set as reference views. They are referred to as \textit{\FTThree} and \textit{\FTAll}, respectively.

\subsection{Initialize 3D Scene from Scratch}
\label{subsec:method_scratch}
While many 3D inpainting methods~\cite{3dgic, gaussian_grouping} only run finetuning methods, we introduce a novel approach that initializes a scene from scratch.
We find that restarting from scratch produces higher quality results than the comparable finetuning approach. We refer to this method as~\textit{\Init} with the 2D inpainters LaMa~\cite{LaMa} and PowerPaint~\cite{powerpaint}.

It is vital that this approach inpaints the images after object removal, since details behind the object can be known from different views. To underline this statement, we introduce a straightforward baseline method using Google's Nano Banana~\cite{nanobanana}, also known as Gemini~2.5 Flash Image. Utilizing Nano Banana~\cite{nanobanana}, we inpaint the object in all input images of the scene in 2D and use the results to render the 3D scene with Gaussian Grouping~\cite{gaussian_grouping}. In the following sections, this approach is referred to as \textit{\InitNanoBanana}.

%% file: sec/4_experiments.tex
\section{Experiments}
\label{sec:exp}
\subsection{Scenes}
The bear scene is taken from the InNeRF360 dataset~\cite{innerf360}, and the kitchen scene is taken from the Mip-NeRF 360 dataset~\cite{mipnerf360}. Additionally, we introduce a novel living room scene with multiple objects. The to-be-removed object, a white plush, is noticeably larger than surrounding objects to ensure that it completely covers them from many views, creating a challenging occlusion scenario. To compensate for the lack of real-world ground truth data in the 360° setting, we also record the scene without the white plush.

\subsection{2D Inpainter Evaluation}
\label{subsec:inpainter_eval}
In the bear scene in~\cref{fig:inpainter_comparison}, LaMa~\cite{LaMa} creates a smooth surface in the masked area with fewer details than PowerPaint~\cite{powerpaint} and Nano Banana~\cite{nanobanana}.
\input{full_figures/2D_inpainter_comparison}
A clean inpainting mask is vital for inpainting with LaMa~\cite{LaMa} and PowerPaint~\cite{powerpaint}. When the mask is too small, they~\cite{LaMa, powerpaint} struggle to remove the entire hole in the kitchen scene. A too large mask results in omitting additional objects in our living room scene.
Nano Banana~\cite{nanobanana} inpaints the object rather than the hole. This leads to the black box behind the object being unseen in the input image and disappearing completely in our living room scene.
BrushNet~\cite{brushnet} produces very poor results in~2D and is therefore not evaluated in~3D.

\subsection{Quantitative Evaluation}
\label{subsec:quan_eval}
We quantitatively evaluate our approaches in~\cref{tab:quan_eval} with the metrics LPIPS~\cite{LPIPS}, PSNR, FID~\cite{FID}, SSIM~\cite{SSIM}, and the respective masked metrics calculated within the inpainted region.
\input{full_figures/eval_table}
\InitLaMa performs the best across most metrics. Methods using LaMa~\cite{LaMa} almost consistently outperform the ones inpainting with PowerPaint~\cite{powerpaint}. \InitNanoBanana shows poor results in all metrics when considering the entire image instead of just the masked region. However, the results of the masked metrics are very competitive. This underlines the sharp inpainting quality of Nano Banana~\cite{nanobanana} within the inpainting masks, but since the approach inpaints the entire object, the region behind it inaccurately reflects the ground truth.
Independent of the 2D inpainter, \Init outperforms the two finetuning methods. The finetuning results with LaMa~\cite{LaMa} are better when using all views as reference view in comparison to using only three. When using PowerPaint~\cite{powerpaint}, it is the other way around.

\subsection{Qualitative Evaluation}
In \cref{fig:3D_comparison}, we qualitatively evaluate the results of our approaches in the three scenes.
\input{full_figures/3D_comparison}

\textbf{Bear Inpainting Quality and Background Blur}. In the bear scene, the inpainting quality of all methods looks very similar despite the differences in~2D. The masked region in all methods is smooth, which is because the details of the generative models~\cite{nanobanana, powerpaint} differ in each view, leading to them blending together in~3D
\InitNanoBanana produces blurry background artifacts because of hallucinations in unknown regions behind the object.
The remaining six methods have blurry background results because the generated inpainting masks by 3DGIC~\cite{3dgic} produce artifacts in the background above the hole that the respective~2D inpainter fills. Because of the differences in these inpainted regions, blurry results arise.

\textbf{Kitchen Inpainting Quality}. 
The results of the 2D inpainter are accurately reflected in~3D. \InitNanoBanana produces the highest quality results, while the other methods fail to fully remove the hole from the scene.

\textbf{Living Room Multi-Object Handling}. The main issue of \InitNanoBanana of having no 3D awareness when inpainting is particularly evident in our living room scene. 
From a given view, objects behind the removed object are omitted in the 3D setting as shown in \cref{fig:nano_living_room}.

\input{full_figures/nano_living_room}

%% file: full_figures/2D_inpainter_comparison.tex
\begin{figure*}[t]
     \centering
     \begin{subfigure}[b]{0.16\textwidth}
         \centering
         \includegraphics[width=\textwidth]{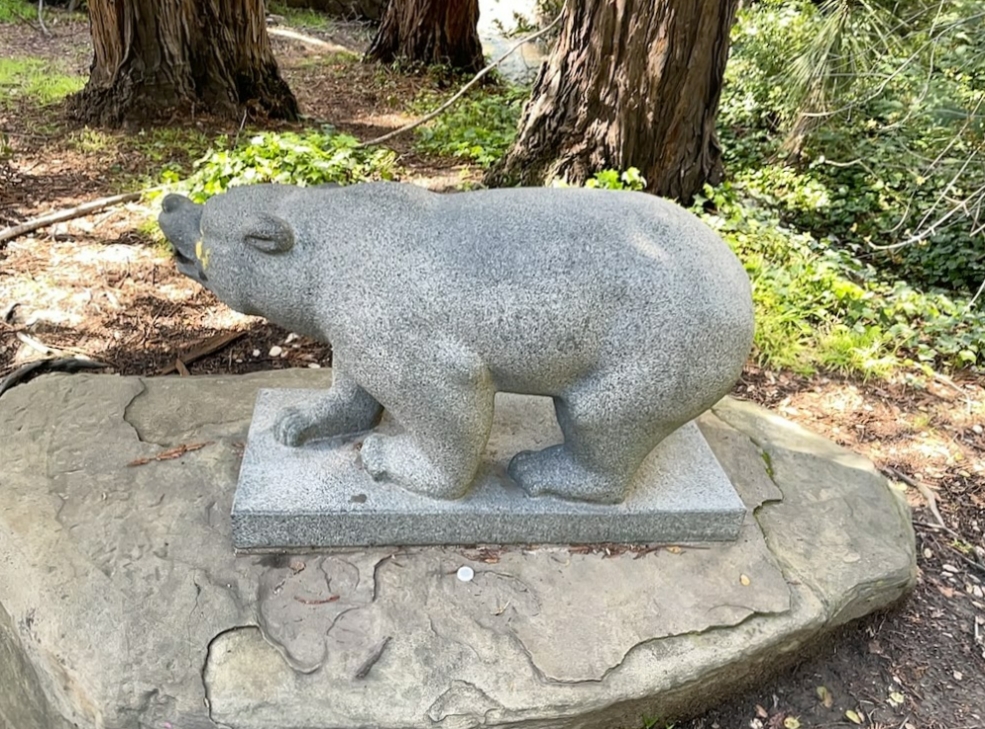}
     \end{subfigure}
     \hfill
     \begin{subfigure}[b]{0.16\textwidth}
         \centering
         \includegraphics[width=\textwidth]{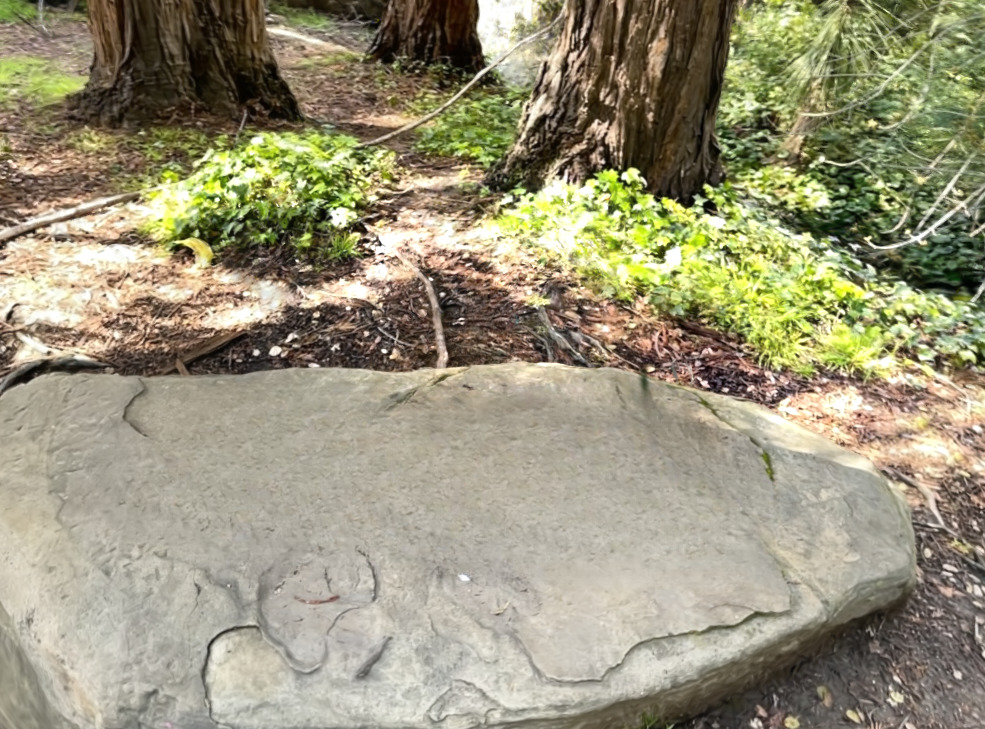}
     \end{subfigure}
     \hfill
     \begin{subfigure}[b]{0.16\textwidth}
         \centering
         \includegraphics[width=\textwidth]{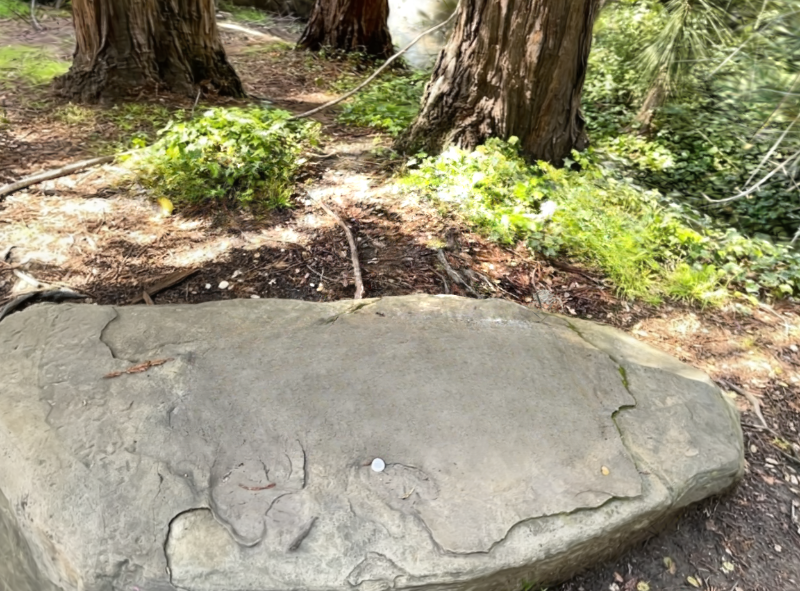}
     \end{subfigure}
      \hfill
     \begin{subfigure}[b]{0.16\textwidth}
         \centering
         \includegraphics[width=\textwidth]{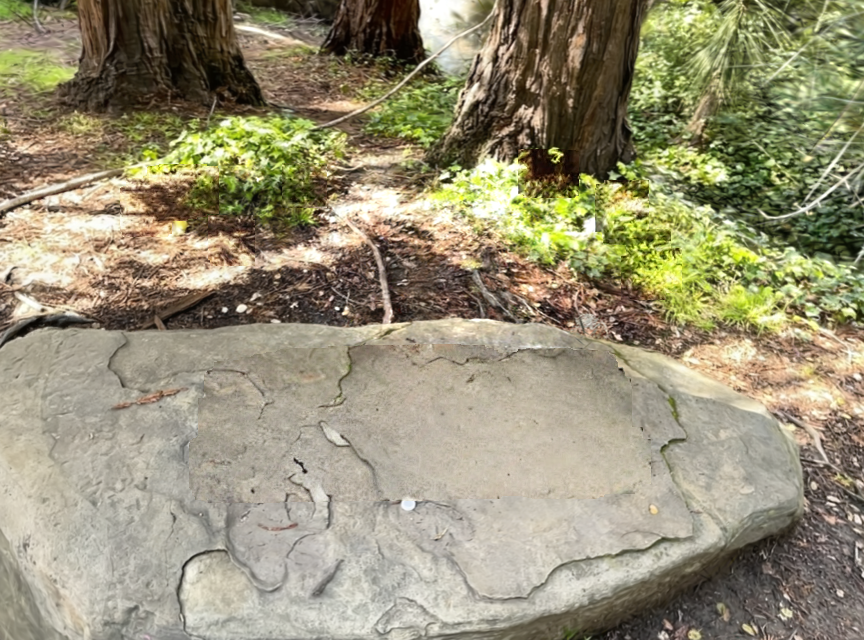}
     \end{subfigure}
      \hfill
     \begin{subfigure}[b]{0.16\textwidth}
         \centering
         \includegraphics[width=\textwidth]{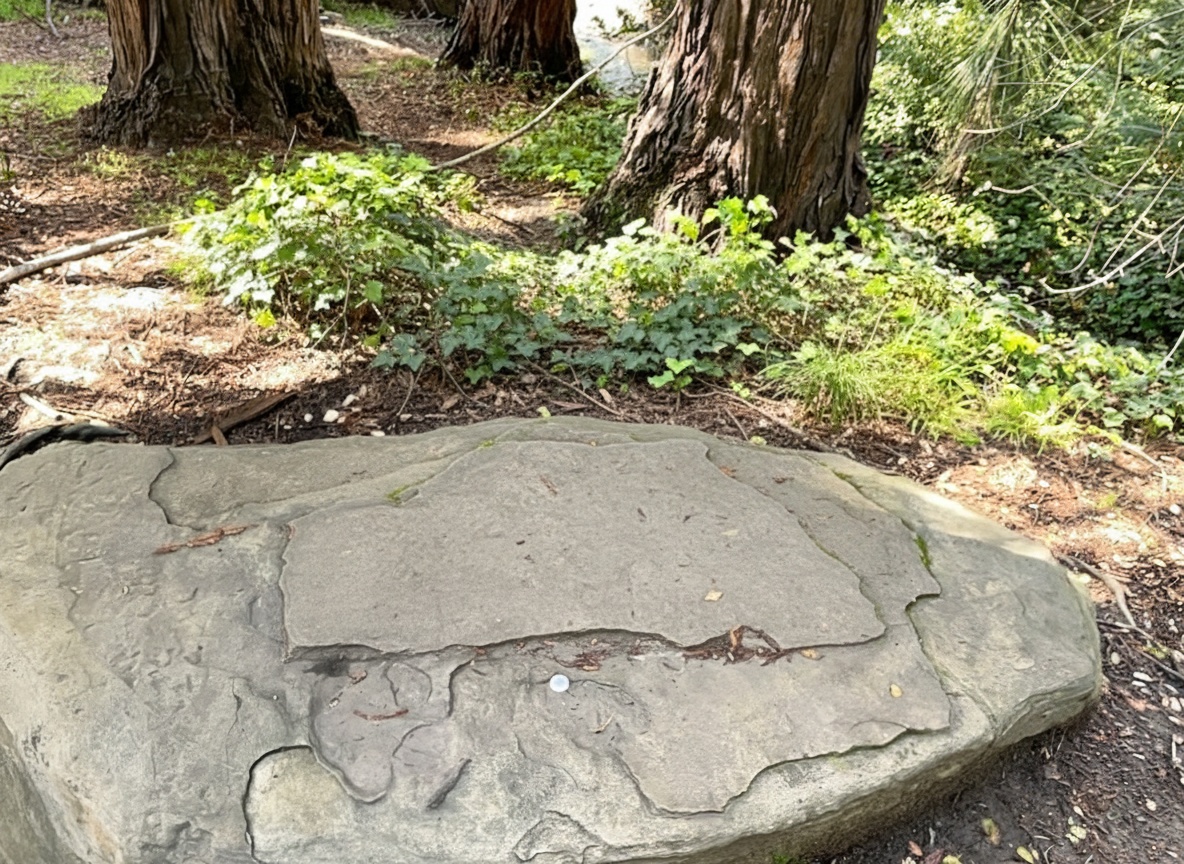}
     \end{subfigure}
      \hfill
     \begin{subfigure}[b]{0.16\textwidth}
         \centering
         \includegraphics[width=\textwidth]{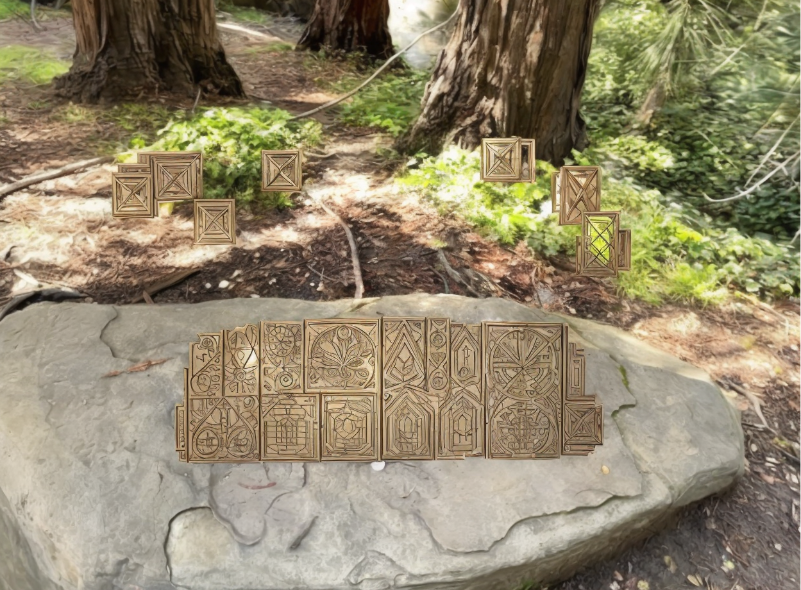}
     \end{subfigure}

     \vspace{0.1cm} 

     \begin{subfigure}[b]{0.16\textwidth}
         \centering
         \includegraphics[width=\textwidth]{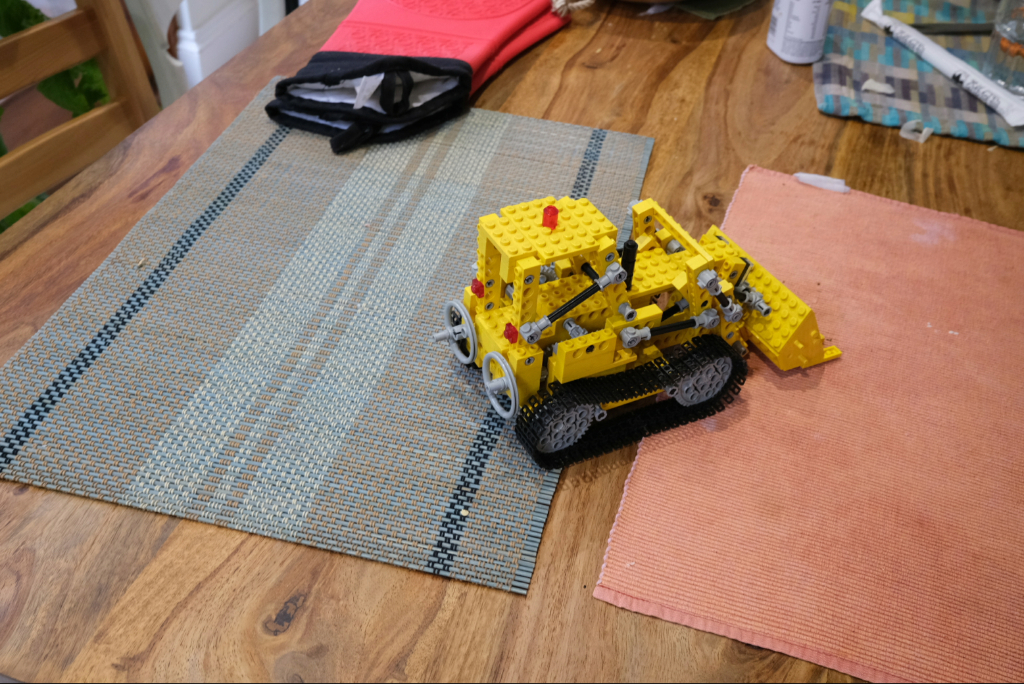}
     \end{subfigure}
     \hfill
     \begin{subfigure}[b]{0.16\textwidth}
         \centering
         \includegraphics[width=\textwidth]{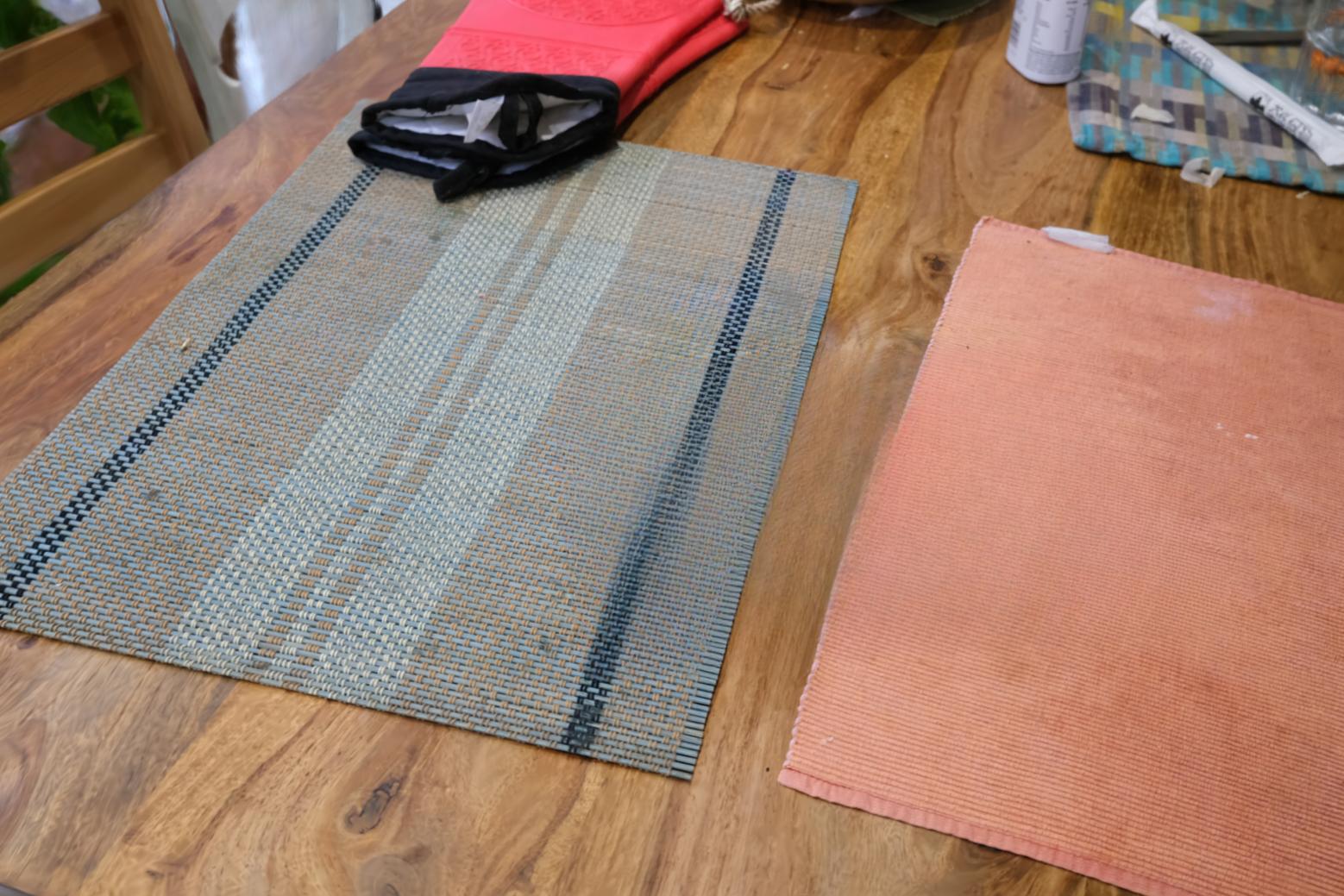}
     \end{subfigure}
     \hfill
     \begin{subfigure}[b]{0.16\textwidth}
         \centering
         \includegraphics[width=\textwidth]{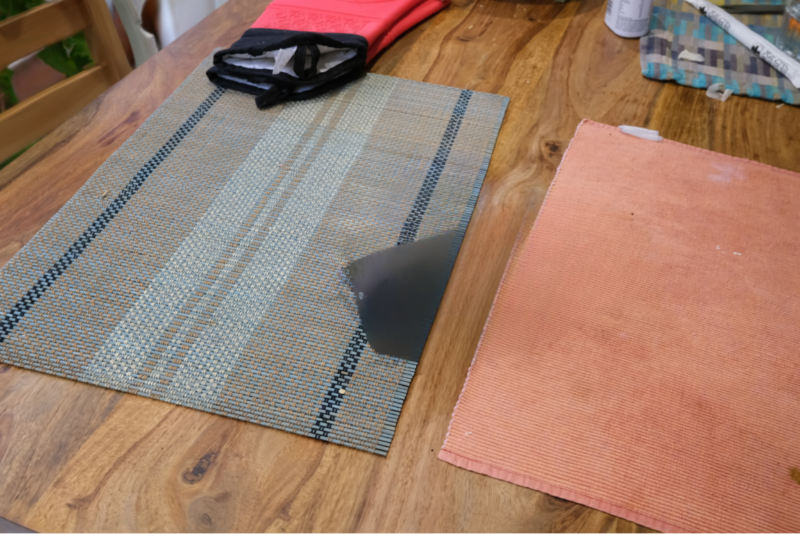}
     \end{subfigure}
      \hfill
     \begin{subfigure}[b]{0.16\textwidth}
         \centering
         \includegraphics[width=\textwidth]{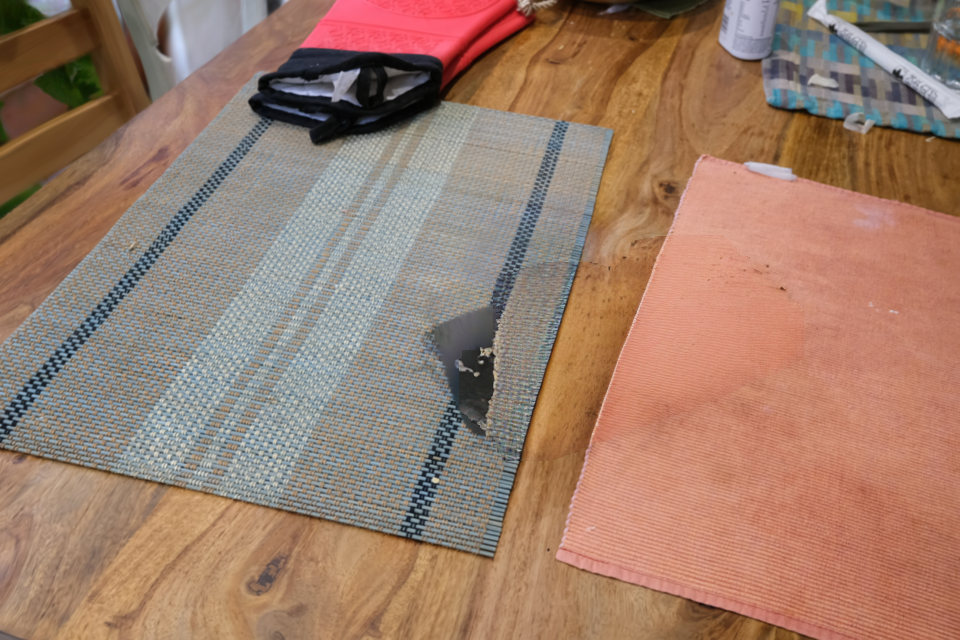}
     \end{subfigure}
      \hfill
     \begin{subfigure}[b]{0.16\textwidth}
         \centering
         \includegraphics[width=\textwidth]{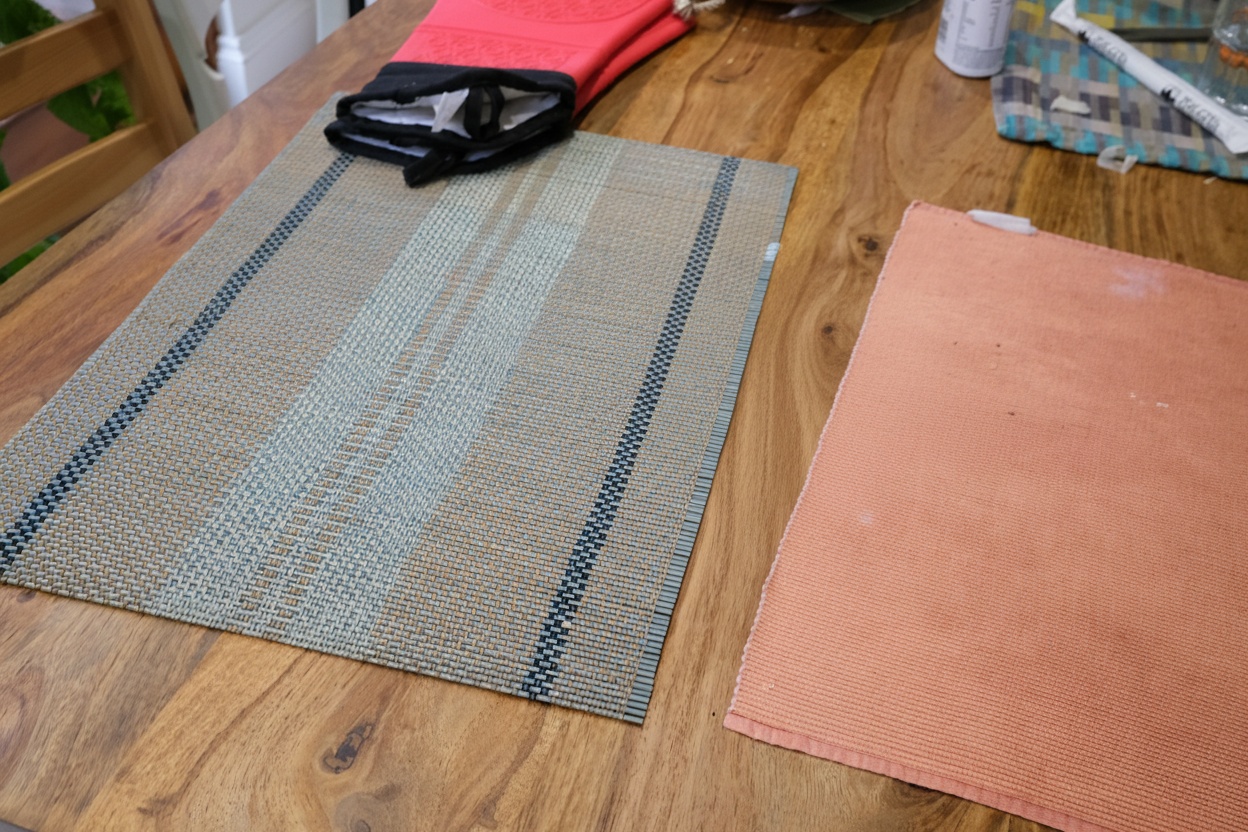}
     \end{subfigure}
      \hfill
     \begin{subfigure}[b]{0.16\textwidth}
         \centering
         \includegraphics[width=\textwidth]{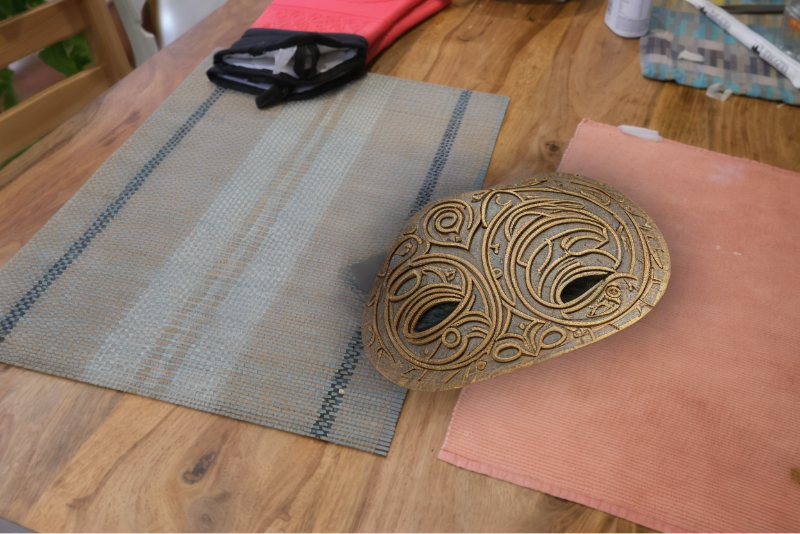}
     \end{subfigure}

     \vspace{0.1cm} 

     \begin{subfigure}[b]{0.16\textwidth}
         \centering
         \includegraphics[width=\textwidth]{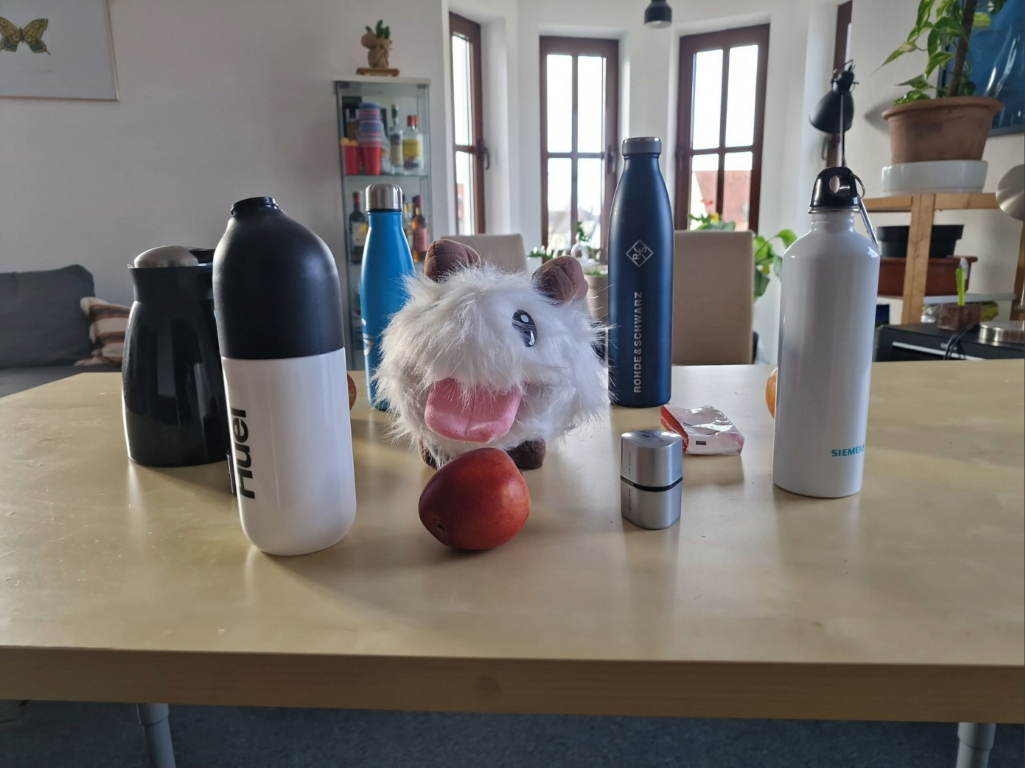}
         \caption{Image with Object}
     \end{subfigure}
     \hfill
     \begin{subfigure}[b]{0.16\textwidth}
         \centering
         \includegraphics[width=\textwidth]{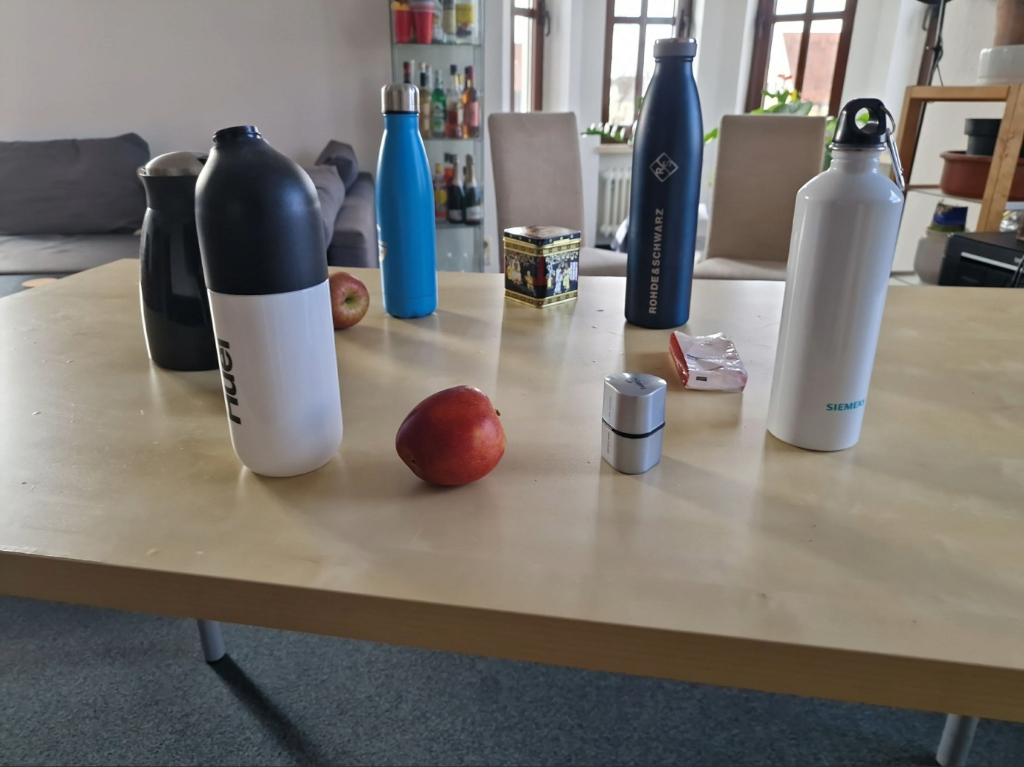}
         \caption{GT}
     \end{subfigure}
     \hfill
     \begin{subfigure}[b]{0.16\textwidth}
         \centering
         \includegraphics[width=\textwidth]{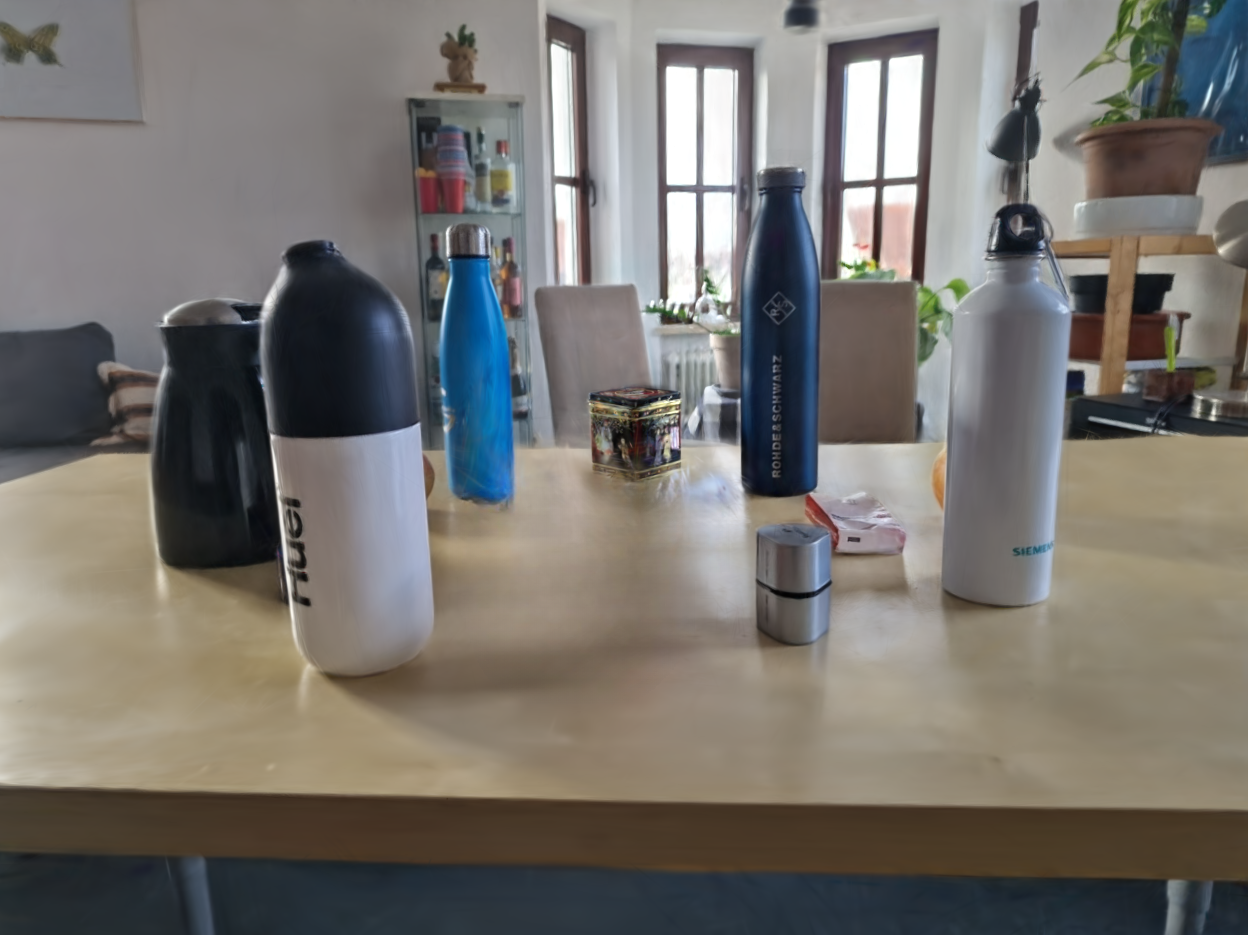}
         \caption{LaMa~\cite{LaMa}}
     \end{subfigure}
      \hfill
     \begin{subfigure}[b]{0.16\textwidth}
         \centering
         \includegraphics[width=\textwidth]{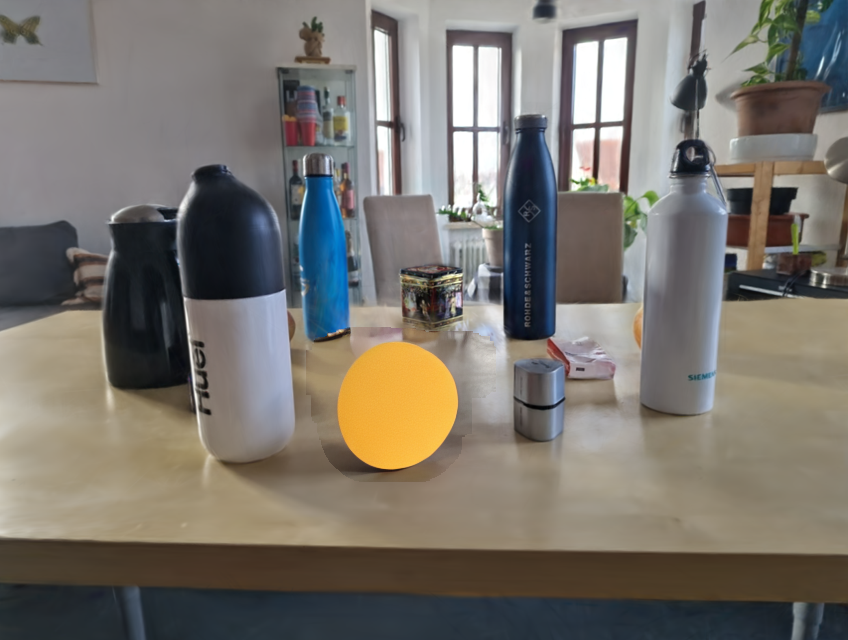}
         \caption{PowerPaint~\cite{powerpaint}}
     \end{subfigure}
      \hfill
     \begin{subfigure}[b]{0.16\textwidth}
         \centering
         \includegraphics[width=\textwidth]{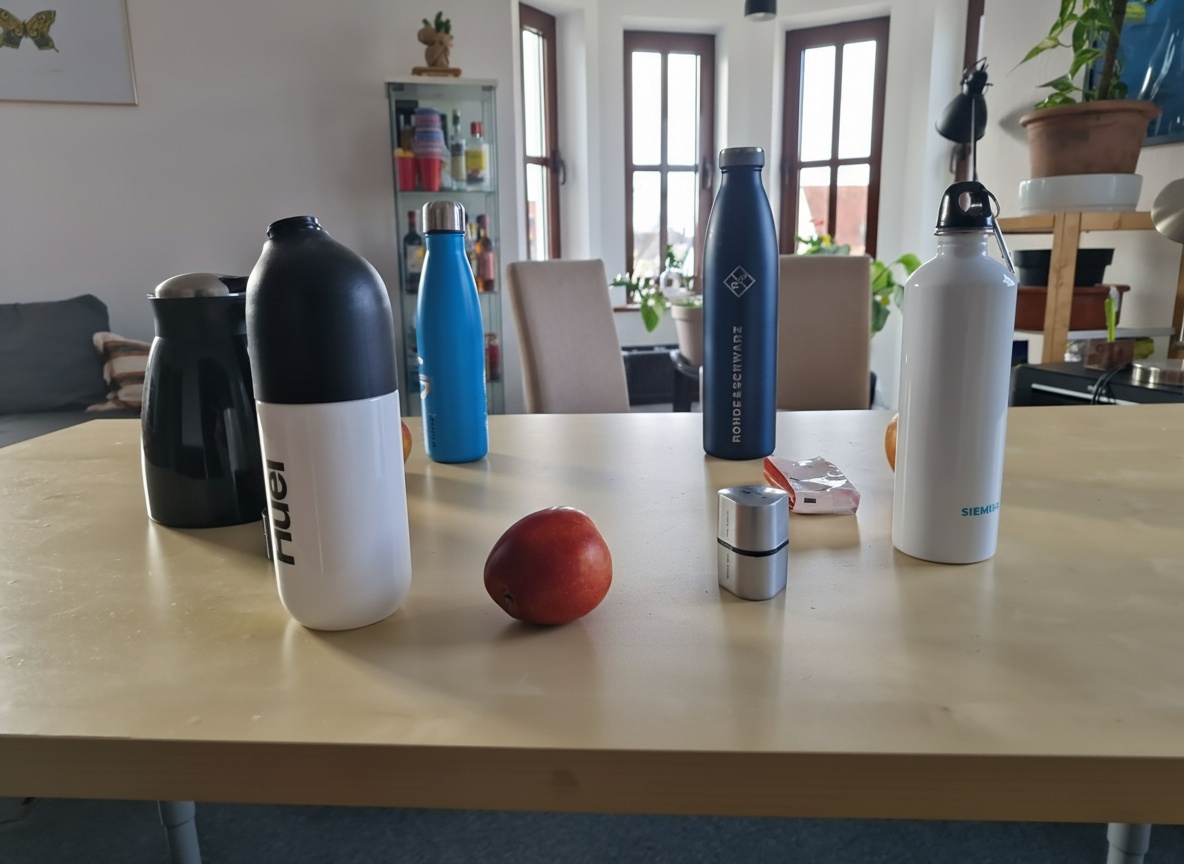}
         \caption{Nano Banana~\cite{nanobanana}}
     \end{subfigure}
      \hfill
     \begin{subfigure}[b]{0.16\textwidth}
         \centering
         \includegraphics[width=\textwidth]{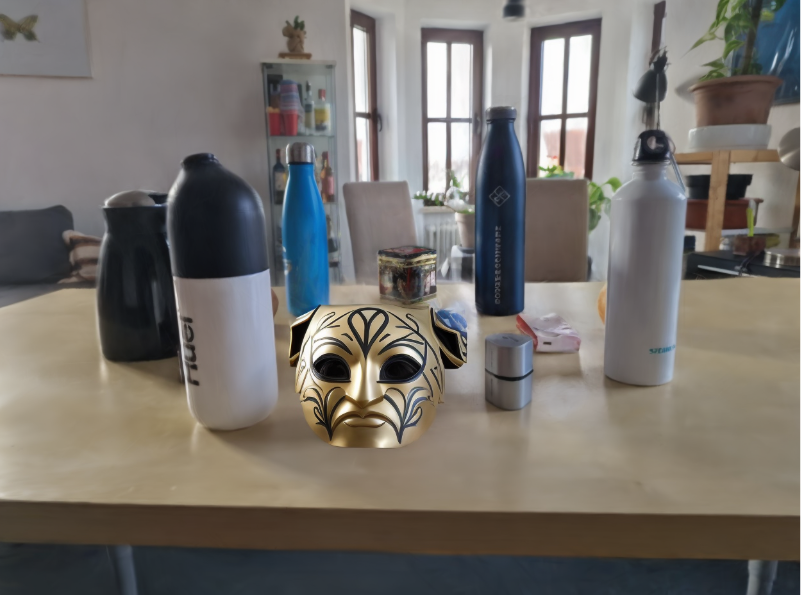}
         \caption{BrushNet~\cite{brushnet}}
     \end{subfigure}

     \caption{\textbf{2D Inpainter Comparison}. We compare the ground truth with the inpainting results from LaMa~\cite{LaMa}, PowerPaint~\cite{powerpaint}, Nano Banana~\cite{nanobanana}, and BrushNet~\cite{brushnet}. We visualize the bear scene (top), the kitchen scene (middle), and our new living room scene (bottom).}
     \label{fig:inpainter_comparison}
\end{figure*}

%% file: full_figures/eval_table.tex
\begin{table*}[t]
    \centering
    \begin{tabular}{@{}llllllllll@{}}
        \toprule
        Method & PSNR $\uparrow$ & m-PSNR $\uparrow$ & SSIM $\uparrow$ & m-SSIM $\uparrow$ & LPIPS $\downarrow$ & m-LPIPS $\downarrow$ & FID $\downarrow$ & m-FID $\downarrow$ & Run time \\ 
        \midrule
        \FTLaMaThreeShort & 17.84 & 18.41 & 0.688 & 0.9377 & 0.4136 & 0.0663 & 126.12 & 182.41 & 4h 22min \\ 
        \FTPowerPaintThreeShort & 16.43 & 17.84 & 0.6647 & 0.9383 & 0.459 & 0.072 & 241.85 & 229.84 & 3h 40min \\ 
        \FTLaMaAllShort & 18.76 & 19.59 & 0.7103 & \textbf{0.9515} & 0.4016 & 0.0588 & 142.88 & 177.67 & 4h 24min \\ 
        \FTPowerPaintAllShort & 15.35 & 18.01 & 0.5575 & 0.9319 & 0.5392 & 0.0761 & 291.77 & 212.7 & 3h 43min \\ 
        \InitLaMaShort & 19.41 & 19.31 & \textbf{0.7301} & 0.9484 & \textbf{0.3767} & \textbf{0.0556} & \textbf{117.88} & \textbf{153.8} & 5h 22min \\ 
        \InitPowerPaintShort & \textbf{19.68} & 19.44 & 0.7279 & 0.9468 & 0.3814 & 0.0589 & 123.57 & 178.62 & 4h 47min \\ 
        \InitNanoBananaShort & 17.88 & \textbf{19.8} & 0.5865 & 0.942 & 0.4408 & 0.0575 & 131.86 & 153.82 & \textbf{1h 27min} \\ 
        \bottomrule
    \end{tabular}
    \caption{\textbf{Quantitative Evaluation of the 3D Inpainting Results}. The masked metrics (e.g., m-PSNR) represent the scores calculated only within the ground truth inpainting mask. The scores are averaged between the three scenes. The best score is highlighted in bold font.}
    \label{tab:quan_eval}
\end{table*}

%% file: full_figures/3D_comparison.tex
\begin{figure*}[t]
     \centering
     \begin{subfigure}[b]{0.12\textwidth}
         \centering
         \includegraphics[width=\textwidth]{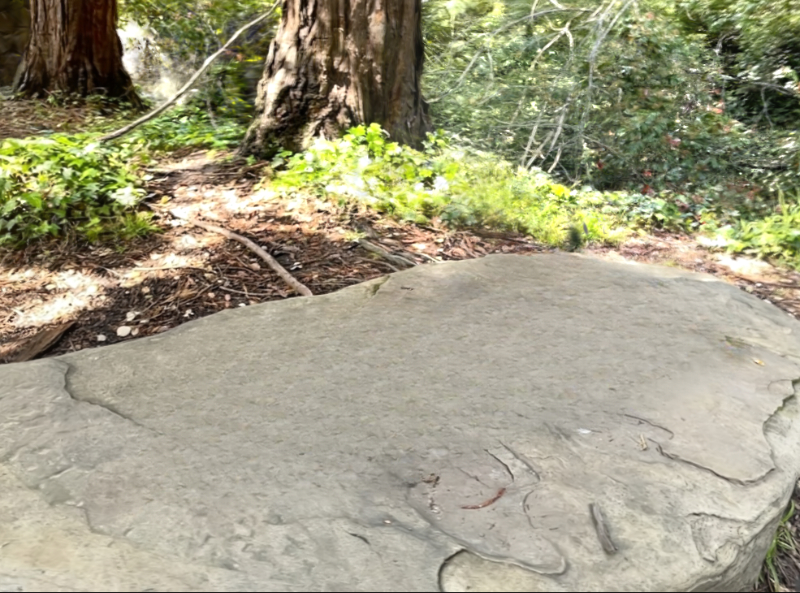}
     \end{subfigure}
      \hfill
     \begin{subfigure}[b]{0.12\textwidth}
         \centering
         \includegraphics[width=\textwidth]{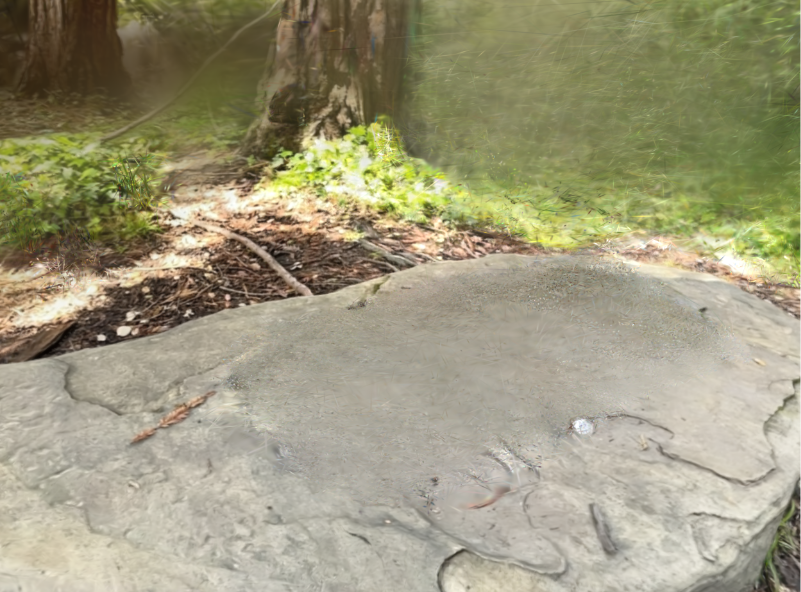}
     \end{subfigure}
     \begin{subfigure}[b]{0.12\textwidth}
         \centering
         \includegraphics[width=\textwidth]{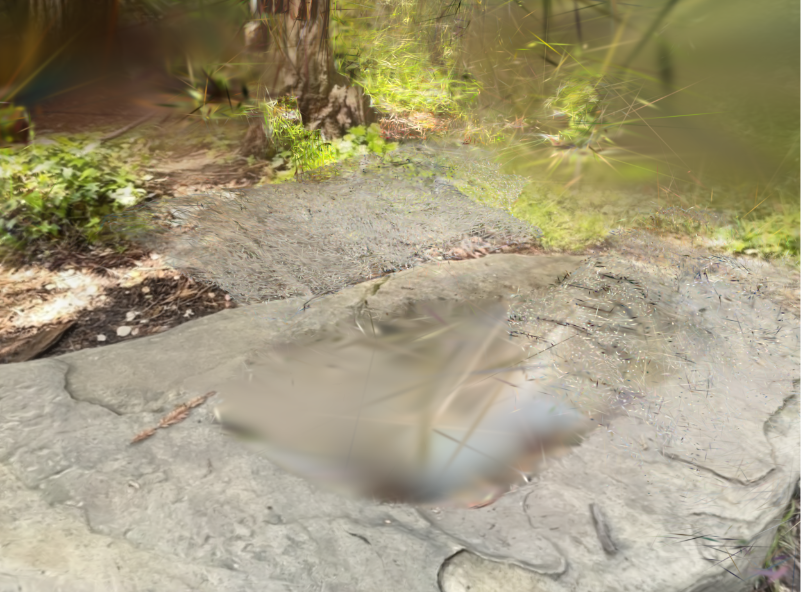}
     \end{subfigure}
     \begin{subfigure}[b]{0.12\textwidth}
         \centering
         \includegraphics[width=\textwidth]{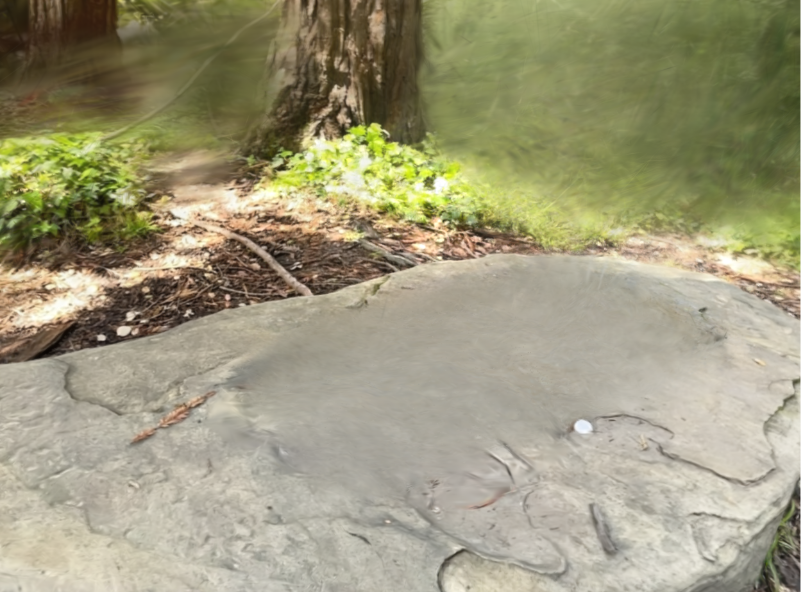}
     \end{subfigure}
     \begin{subfigure}[b]{0.12\textwidth}
         \centering
         \includegraphics[width=\textwidth]{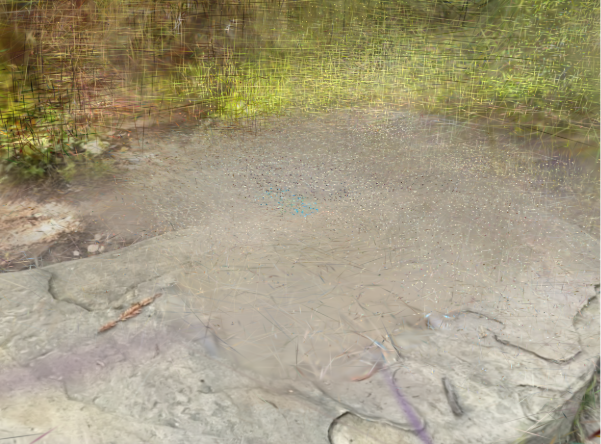}
     \end{subfigure}
     \hfill
     \begin{subfigure}[b]{0.12\textwidth}
         \centering
         \includegraphics[width=\textwidth]{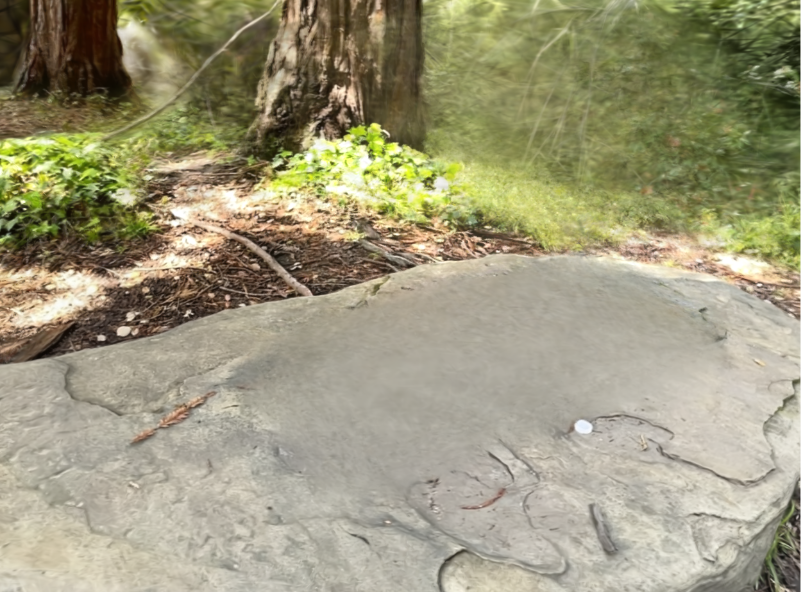}
     \end{subfigure}
      \hfill
     \begin{subfigure}[b]{0.12\textwidth}
         \centering
         \includegraphics[width=\textwidth]{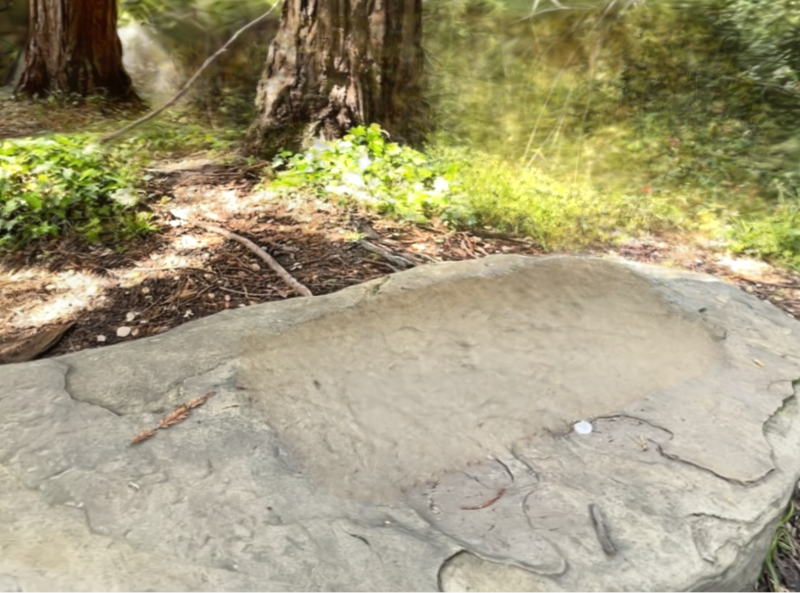}
     \end{subfigure}
     \hfill
     \begin{subfigure}[b]{0.12\textwidth}
         \centering
         \includegraphics[width=\textwidth]{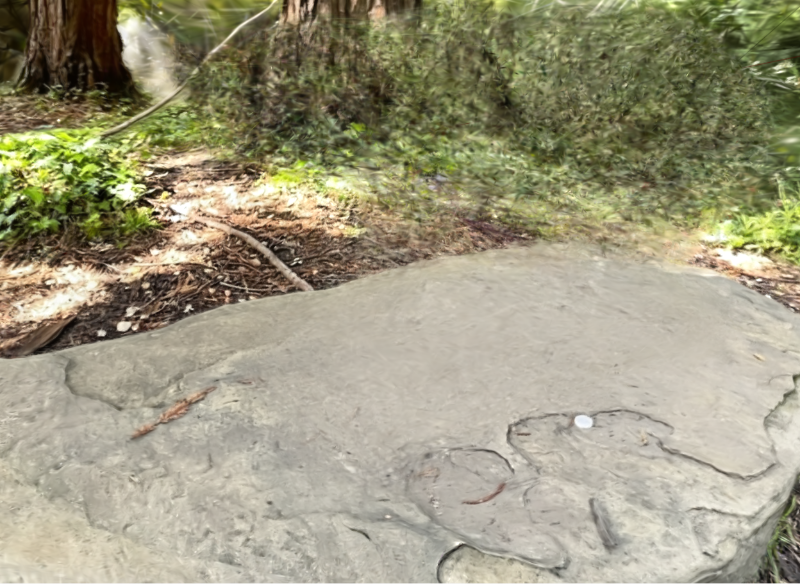}
     \end{subfigure}

     \vspace{0.1cm} 

     \begin{subfigure}[b]{0.12\textwidth}
         \centering
         \includegraphics[width=\textwidth]{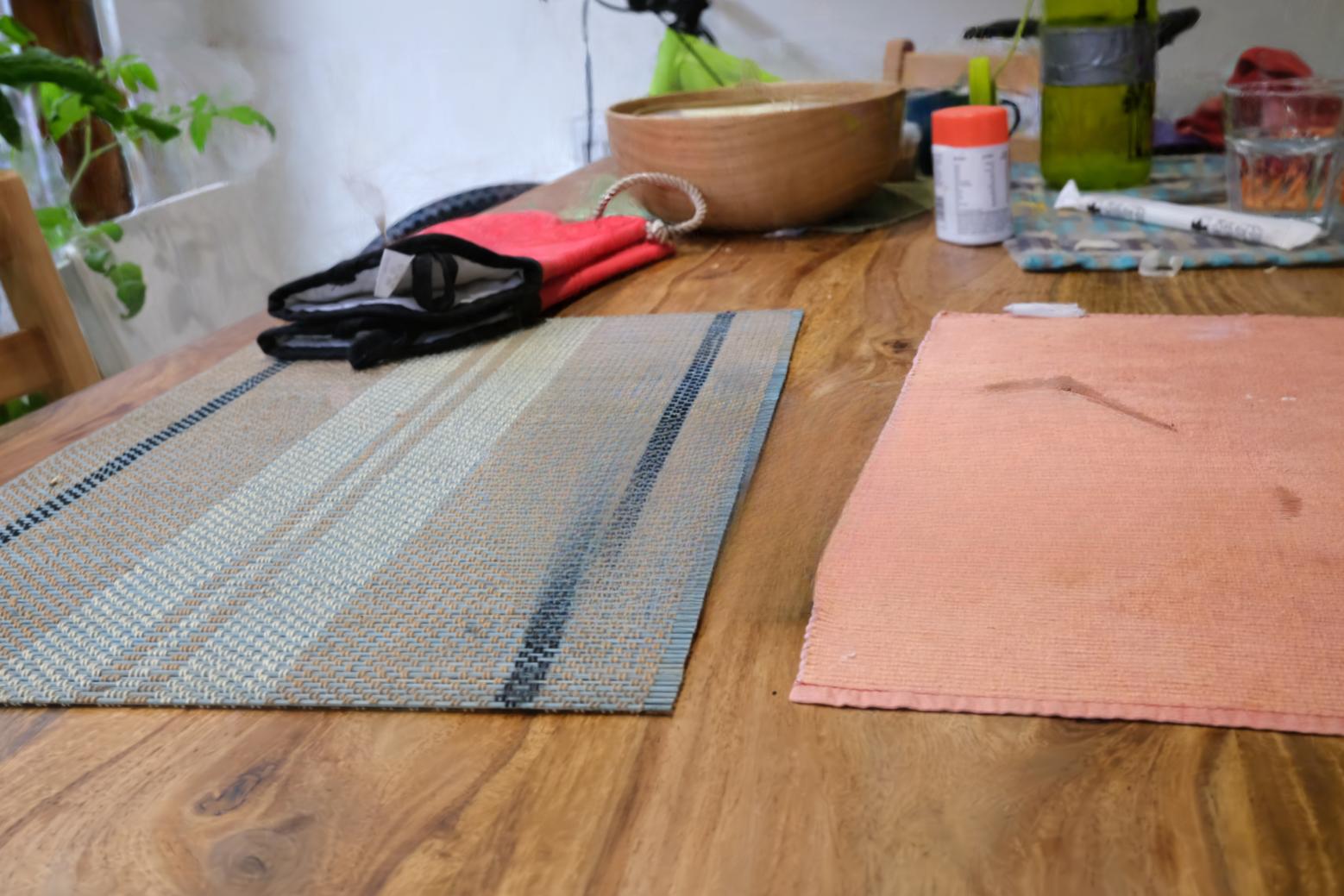}
     \end{subfigure}
     \hfill
     \begin{subfigure}[b]{0.12\textwidth}
         \centering
         \includegraphics[width=\textwidth]{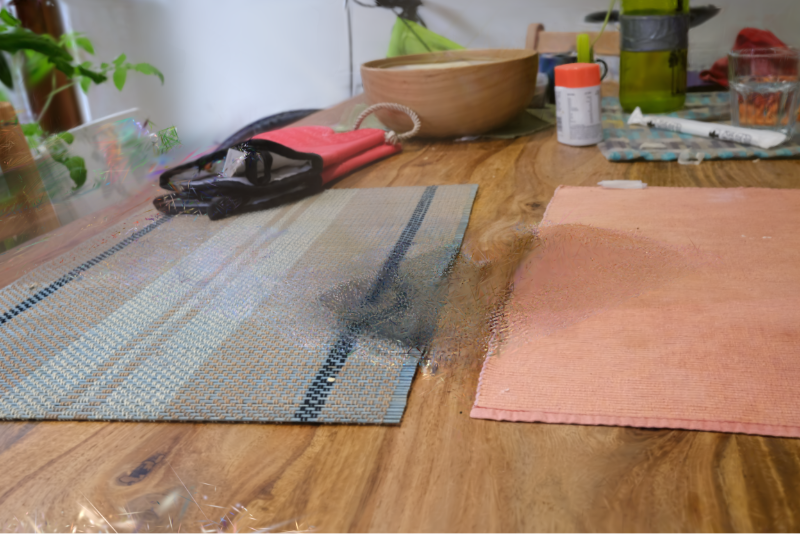}
     \end{subfigure}
     \begin{subfigure}[b]{0.12\textwidth}
         \centering
         \includegraphics[width=\textwidth]{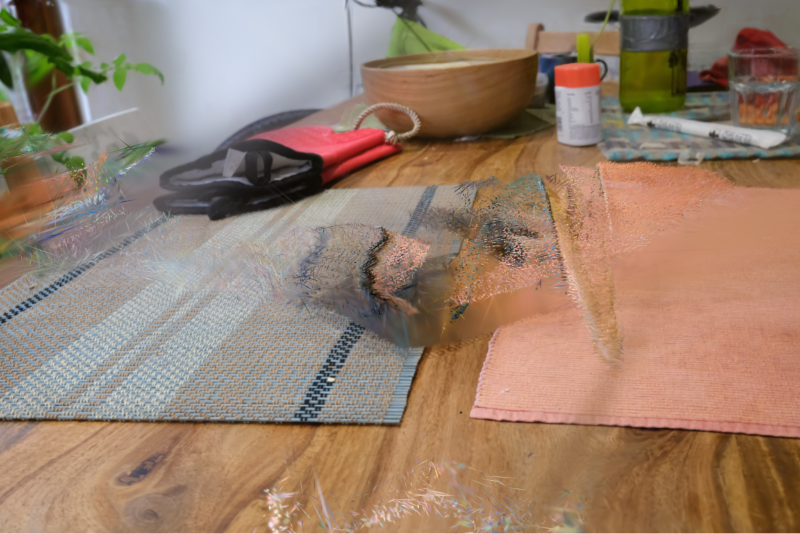}
     \end{subfigure}
     \begin{subfigure}[b]{0.12\textwidth}
         \centering
         \includegraphics[width=\textwidth]{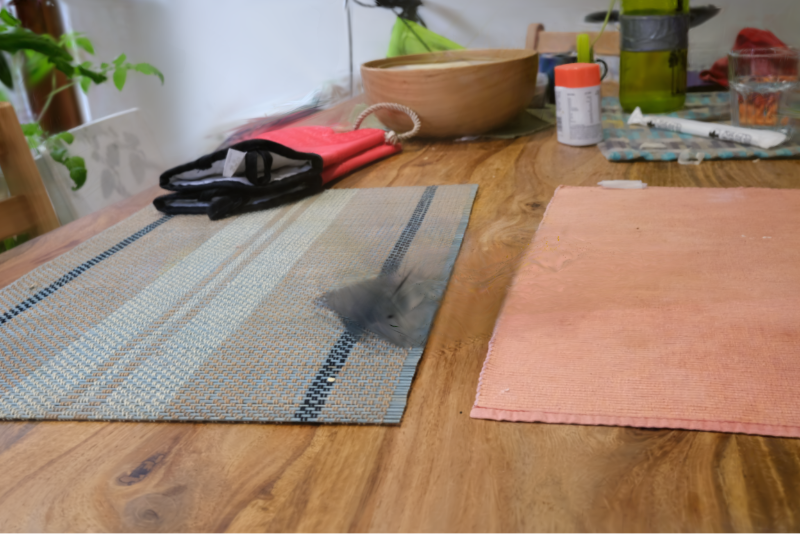}
     \end{subfigure}
     \begin{subfigure}[b]{0.12\textwidth}
         \centering
         \includegraphics[width=\textwidth]{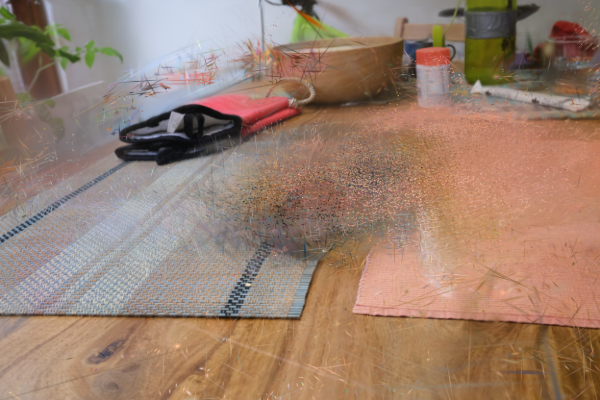}
     \end{subfigure}
     \hfill
     \begin{subfigure}[b]{0.12\textwidth}
         \centering
         \includegraphics[width=\textwidth]{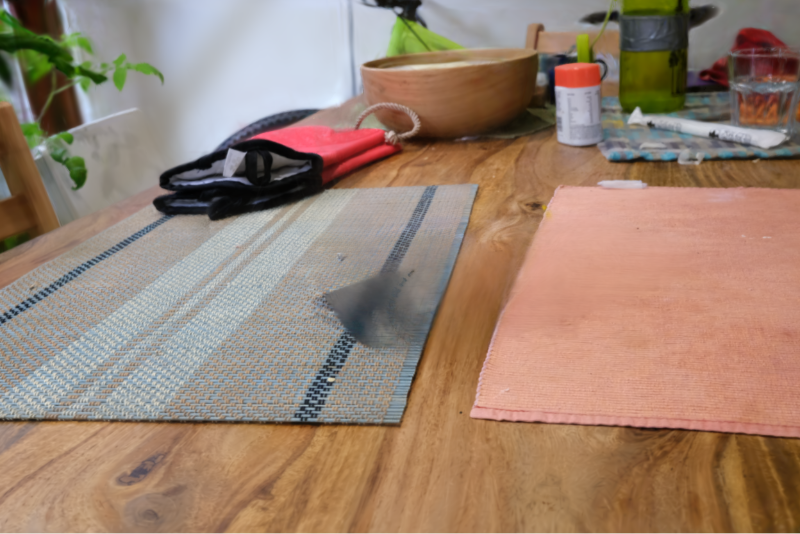}
     \end{subfigure}
      \hfill
     \begin{subfigure}[b]{0.12\textwidth}
         \centering
         \includegraphics[width=\textwidth]{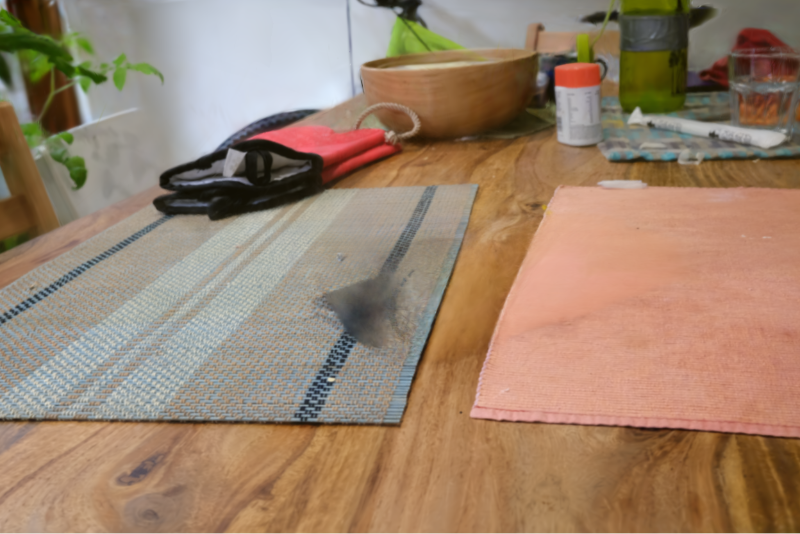}
     \end{subfigure}
      \hfill
     \begin{subfigure}[b]{0.12\textwidth}
         \centering
         \includegraphics[width=\textwidth]{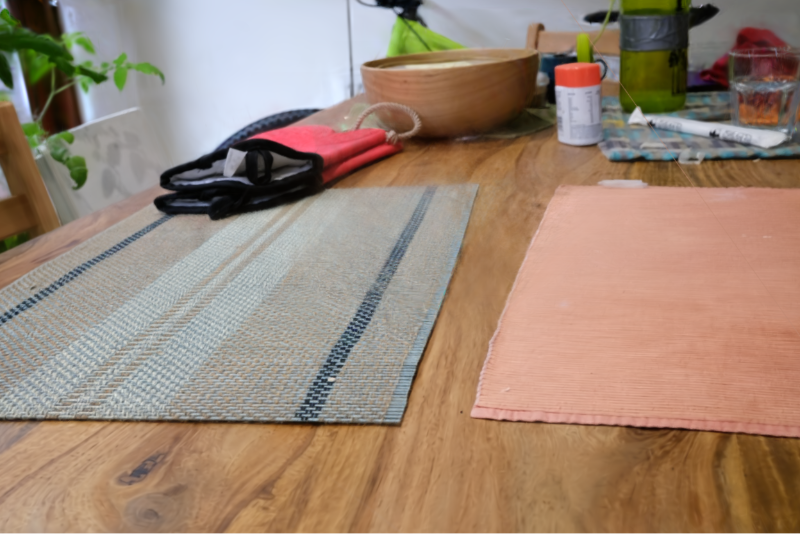}
     \end{subfigure}

     \vspace{0.1cm} 

     \begin{subfigure}[b]{0.12\textwidth}
         \centering
         \includegraphics[width=\textwidth]{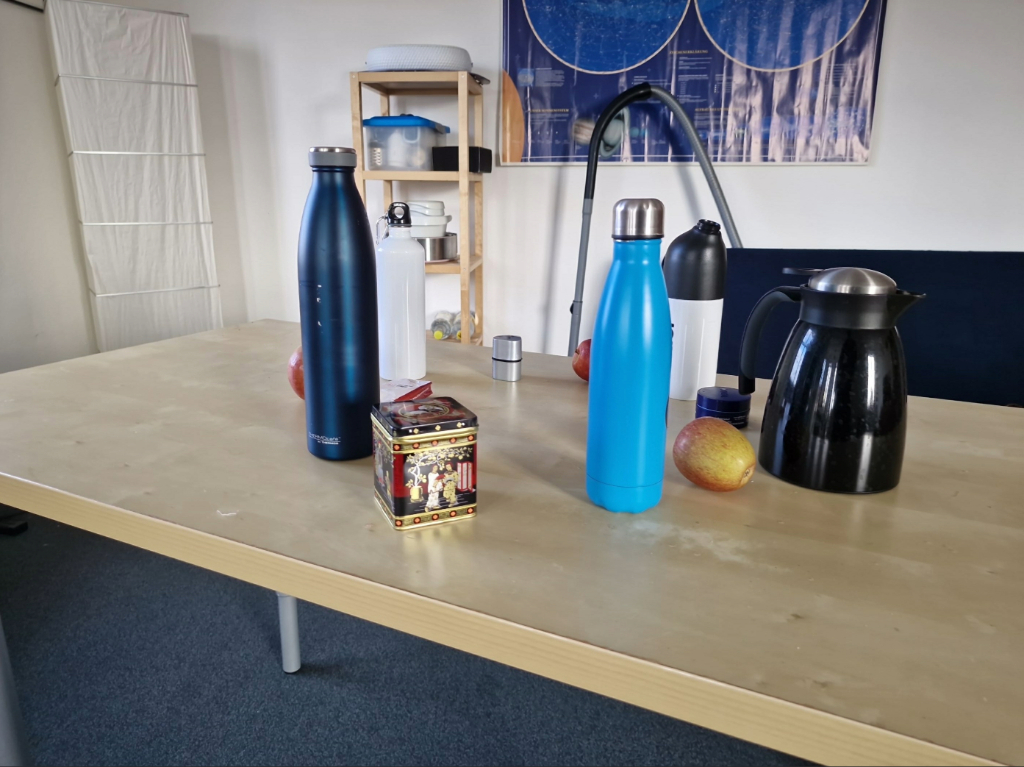}
         \caption{GT}
     \end{subfigure}
      \hfill
     \begin{subfigure}[b]{0.12\textwidth}
         \centering
         \includegraphics[width=\textwidth]{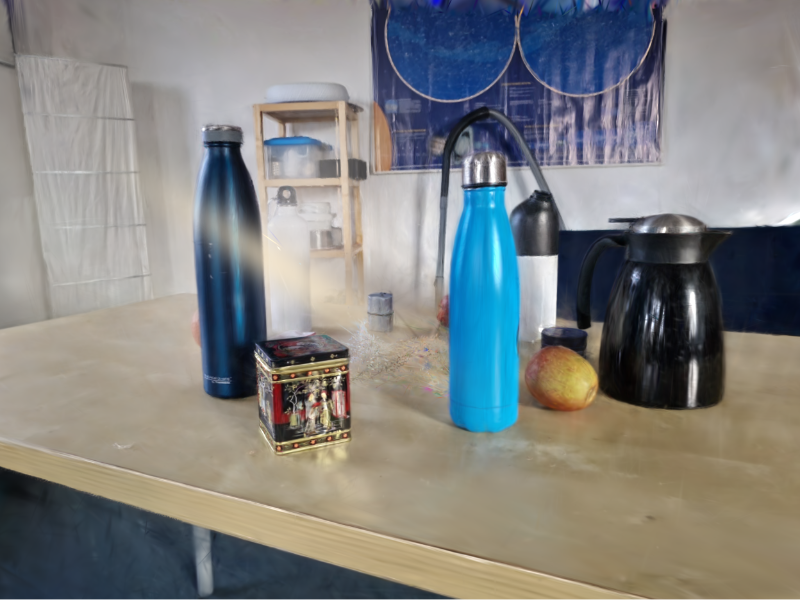}
         \caption{\FTLaMaThreeShort}
     \end{subfigure}
     \begin{subfigure}[b]{0.12\textwidth}
         \centering
         \includegraphics[width=\textwidth]{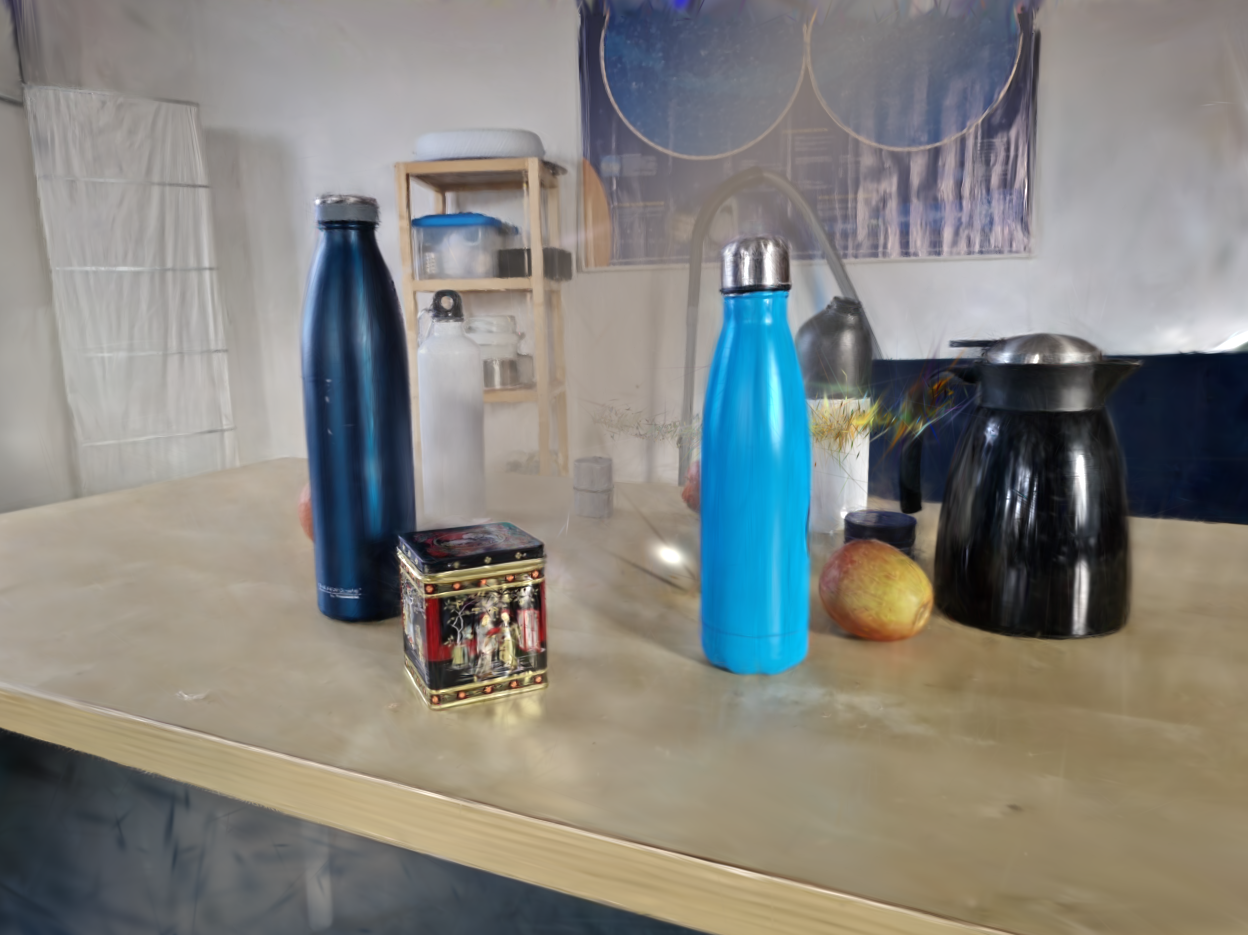}
         \caption{\FTPowerPaintThreeShort}
     \end{subfigure}
     \begin{subfigure}[b]{0.12\textwidth}
         \centering
         \includegraphics[width=\textwidth]{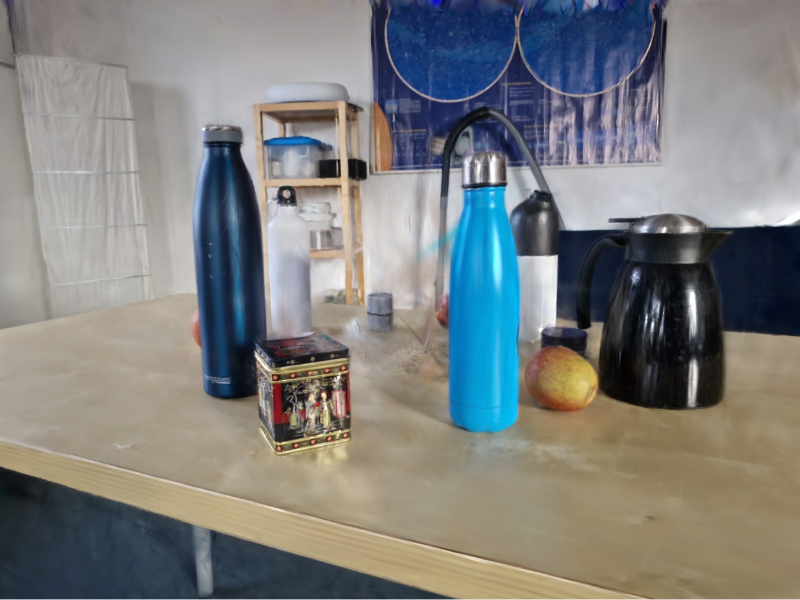}
         \caption{\FTLaMaAllShort}
     \end{subfigure}
     \begin{subfigure}[b]{0.12\textwidth}
         \centering
         \includegraphics[width=\textwidth]{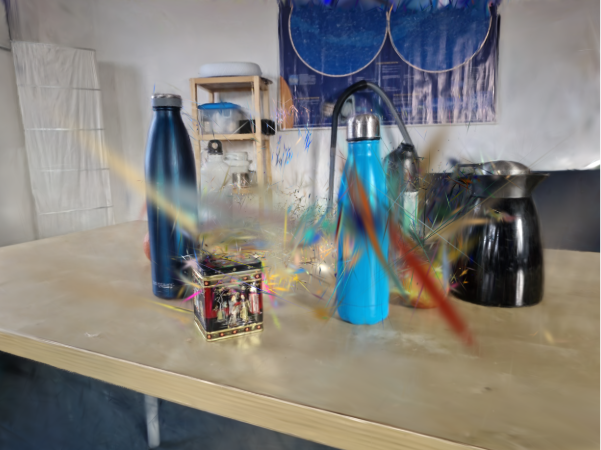}
         \caption{\FTPowerPaintAllShort}
     \end{subfigure}
     \hfill
     \begin{subfigure}[b]{0.12\textwidth}
         \centering
         \includegraphics[width=\textwidth]{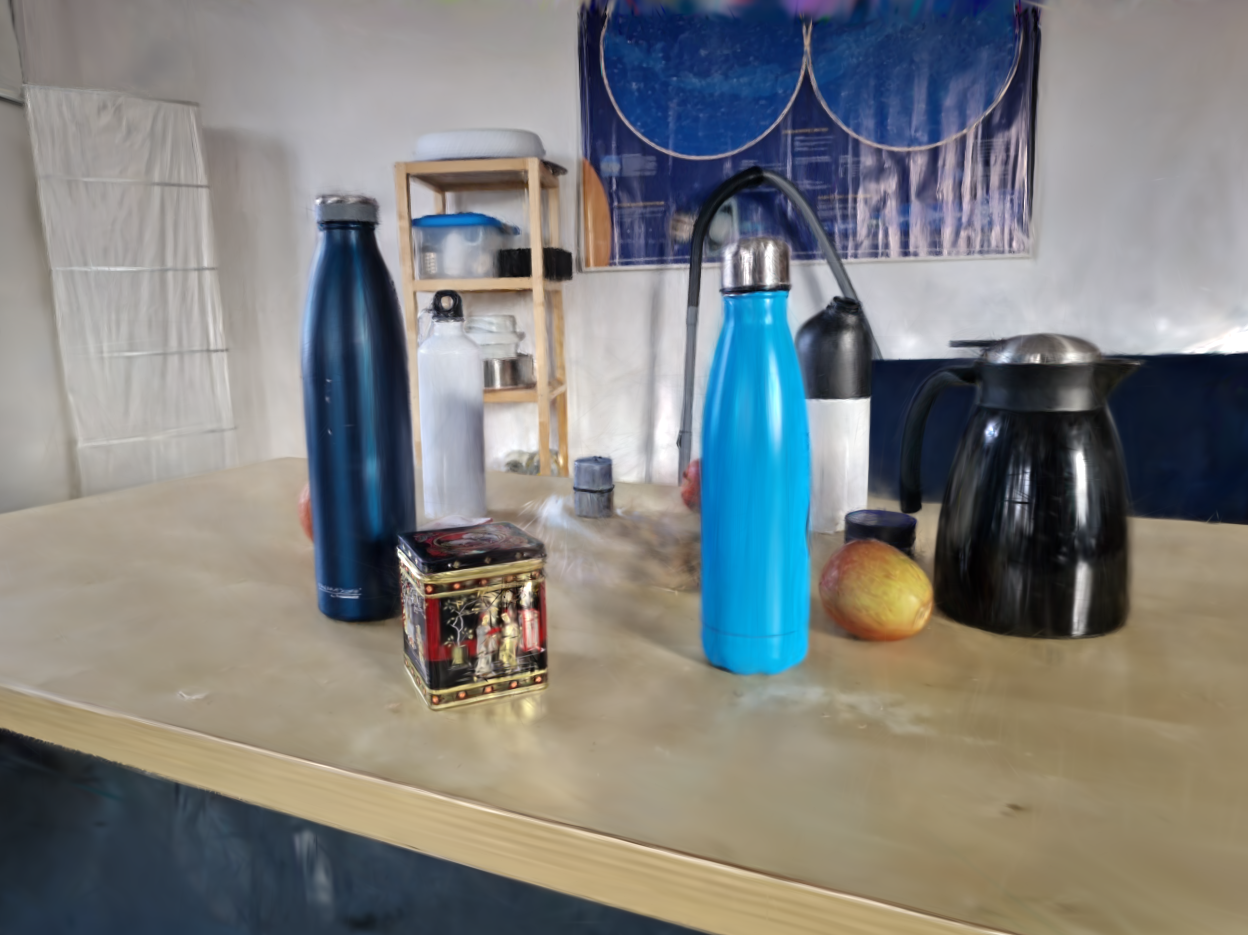}
         \caption{\InitLaMaShort}
     \end{subfigure}
      \hfill
     \begin{subfigure}[b]{0.12\textwidth}
         \centering
         \includegraphics[width=\textwidth]{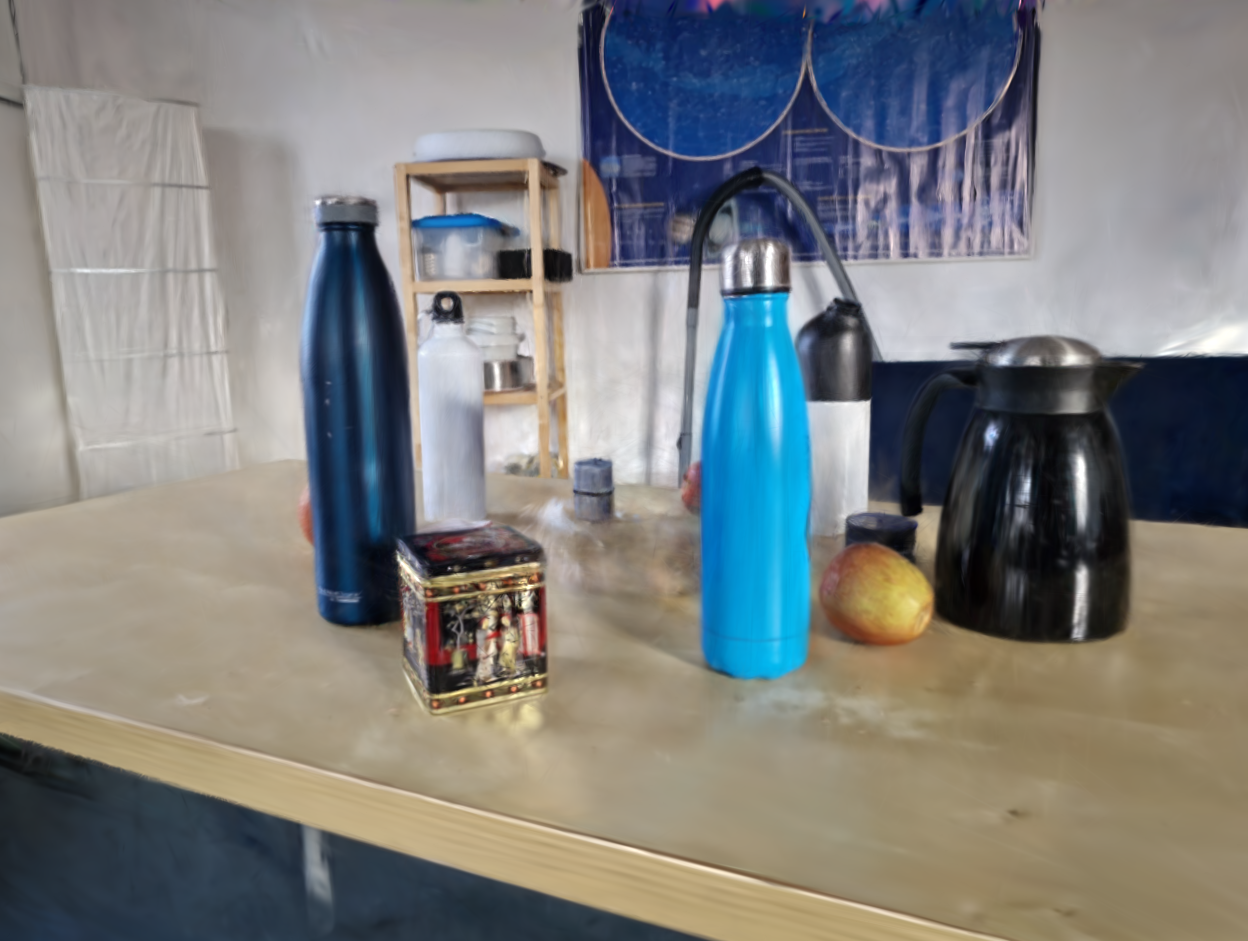}
         \caption{\InitPowerPaintShort}
     \end{subfigure}
     \hfill
     \begin{subfigure}[b]{0.12\textwidth}
         \centering
         \includegraphics[width=\textwidth]{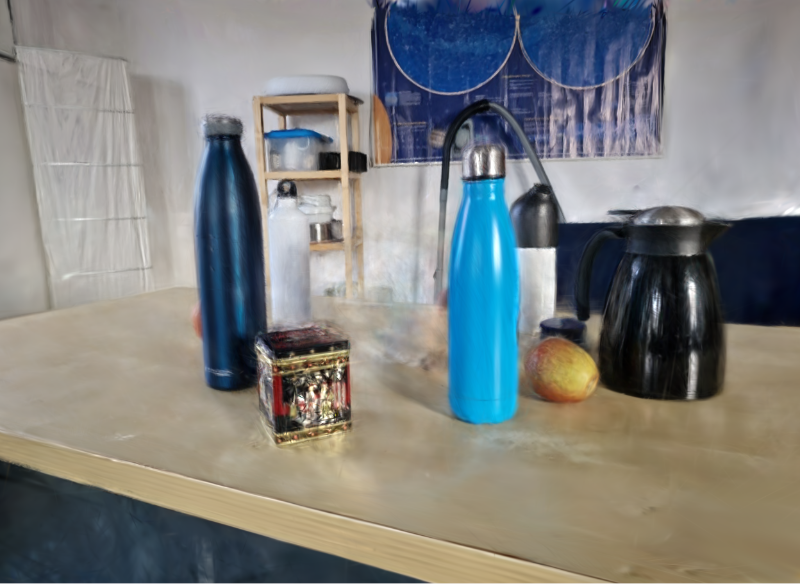}
         \caption{\InitNanoBananaShort}
     \end{subfigure}

     \caption{\textbf{Qualitative Comparison of the 3D Inpainting Results}. We compare the ground truth to our seven different approaches on the bear scene (top), the kitchen scene (middle), and our new living room scene (bottom). The views are all test views of the respective scene.}
     \label{fig:3D_comparison}
\end{figure*}

%% file: full_figures/nano_living_room.tex
\begin{figure}[t]
     \centering
     \begin{subfigure}[b]{0.23\textwidth}
         \centering
         \includegraphics[width=\textwidth]{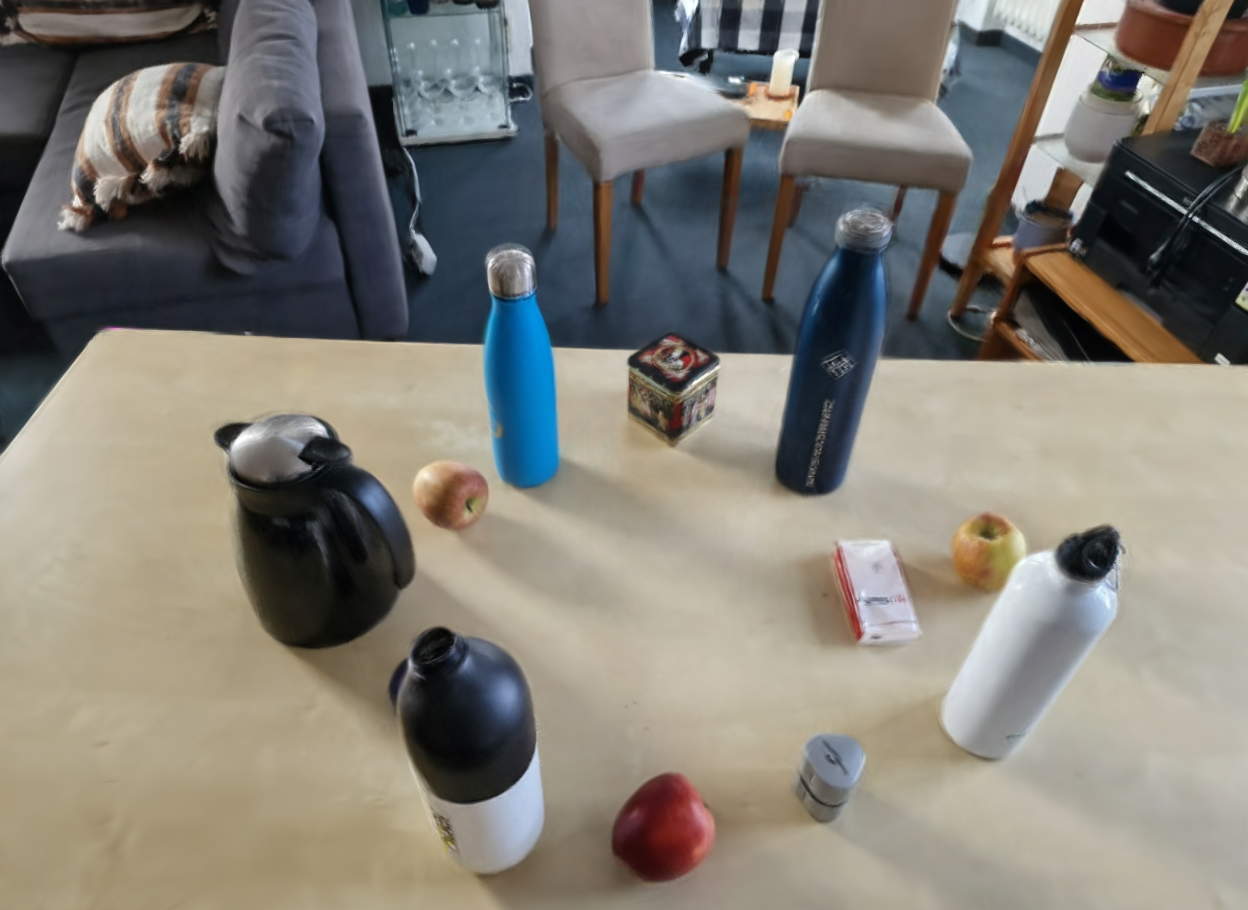}
     \end{subfigure}
      \hfill
     \begin{subfigure}[b]{0.23\textwidth}
         \centering
         \includegraphics[width=\textwidth]{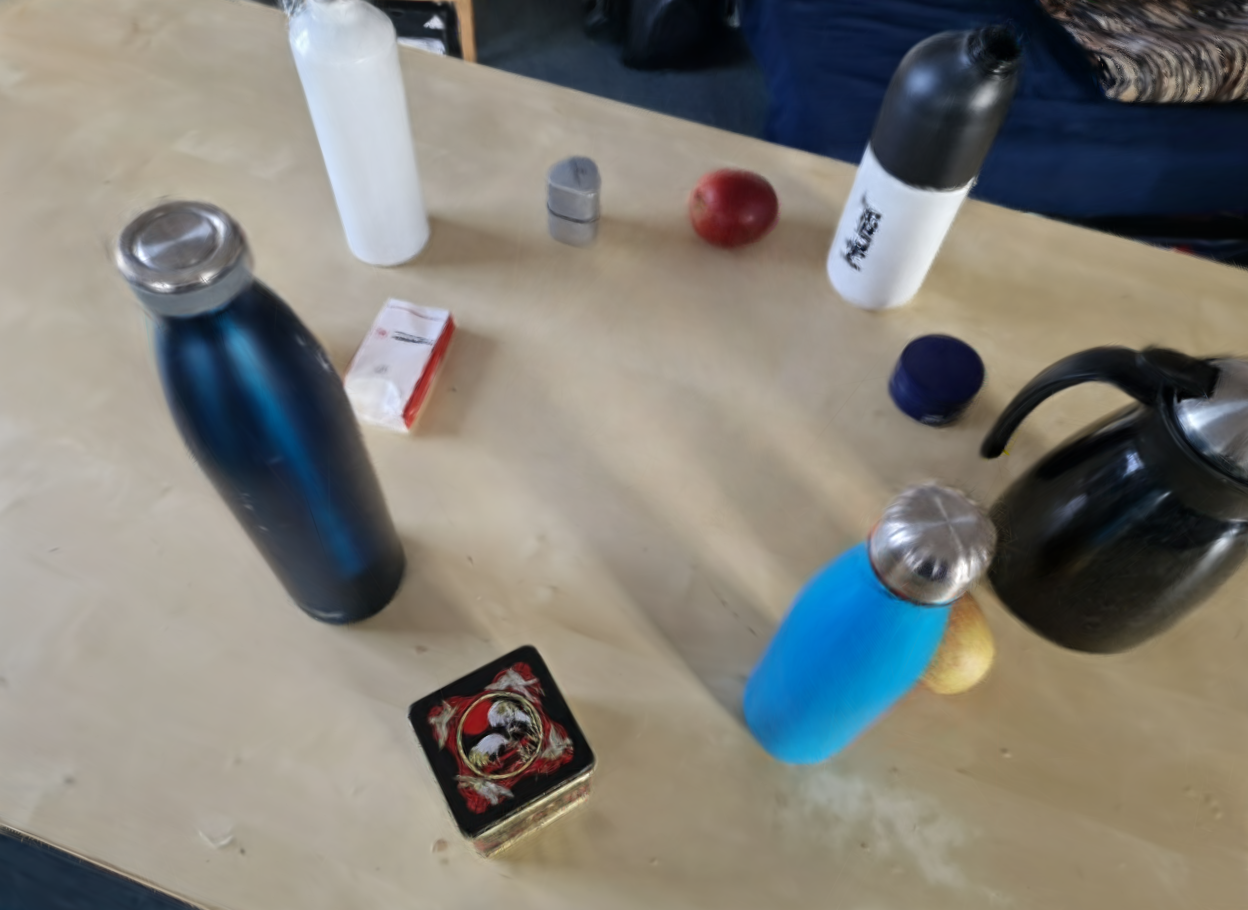}
     \end{subfigure}

     \vspace{0.1cm} 

     \begin{subfigure}[b]{0.23\textwidth}
         \centering
         \includegraphics[width=\textwidth]{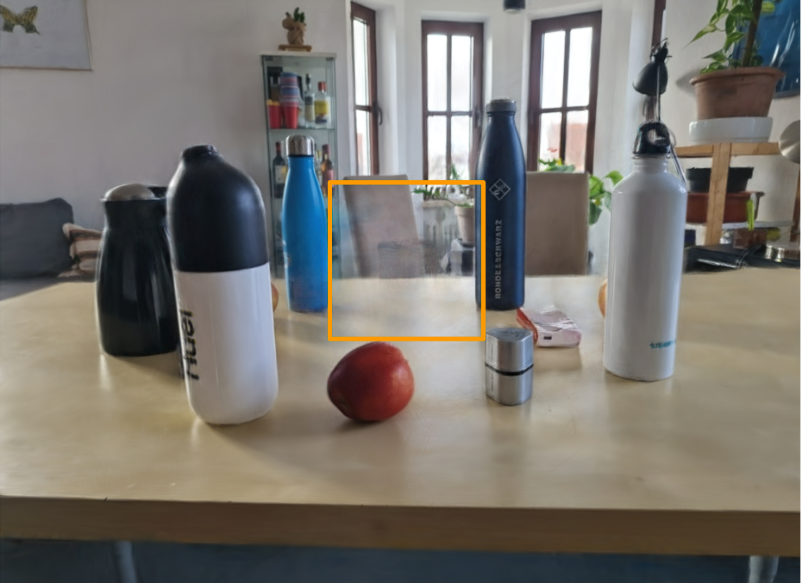}
         \caption{Black box}
     \end{subfigure}
      \hfill
     \begin{subfigure}[b]{0.23\textwidth}
         \centering
         \includegraphics[width=\textwidth]{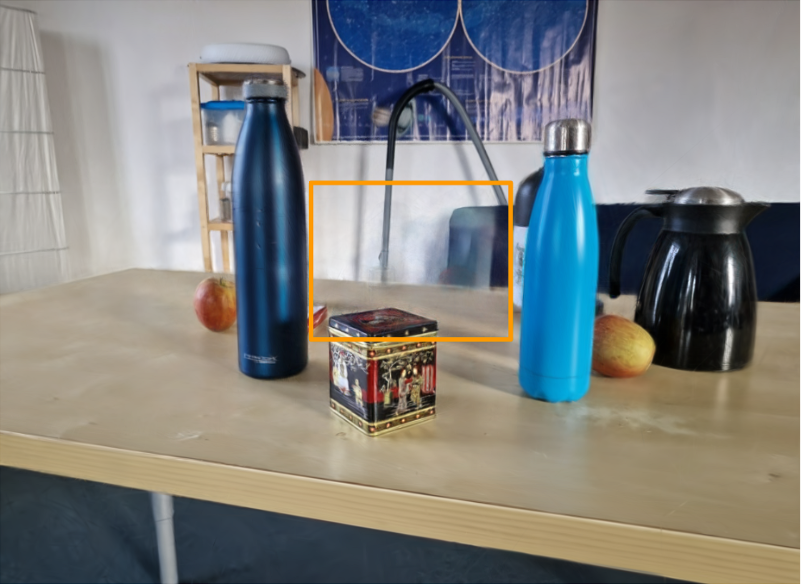}
         \caption{Apple and gray pencil sharpener}
     \end{subfigure}

     \caption{\textbf{Difficult Occlusion Scenario.} All images are 3D results of \InitNanoBanana. 
     All objects are visible in views from above (top) but disappear if they are covered in the input image (bottom).}
     \label{fig:nano_living_room}
\end{figure}

%% file: sec/5_conclusion.tex
\section{Conclusion}
\label{sec:conc}
In this work, we compare different 3D inpainting methods and find that initializing a scene from scratch with the inpainted views qualitatively and quantitatively outperforms comparable finetuning approaches. By evaluating different 2D inpainters~\cite{LaMa, nanobanana, powerpaint, brushnet}, we conclude that LaMa~\cite{LaMa} is better suited for 3D inpainting results in comparison to generative models like PowerPaint~\cite{powerpaint} and BrushNet~\cite{brushnet}. While the generative models~\cite{powerpaint, nanobanana} may produce better results in 2D, LaMa~\cite{LaMa} outperforms them quantitatively and qualitatively in 3D. To compensate for the sparsity of 3D ground truth data in the 360° setting, we introduce a new living room scene with multiple objects. A qualitative analysis of this scene shows the need for object removal before inpainting, as our straightforward \InitNanoBanana approach struggles with details behind the removed object. 

%% file: sec/X_suppl.tex
\clearpage
\setcounter{page}{1}
\maketitlesupplementary

\section{Losses for Finetuning}
For finetuning, we define the following losses.

The loss terms are different for reference views and for non-reference views. We define
\begin{equation}
    \mathcal{L}_{recon} =
    \begin{cases}
        \mathcal{L}^{M}_{1}(I_{in}, I) & \text{if } v \in V_{ref} \\
        \mathcal{L}_{LPIPS}(I_{in}, I) & \text{otherwise,}
    \end{cases}
\end{equation}
where~$V_{ref}$ is the set of reference views,~$\mathcal{L}^{M}_{1}$ is the masked L1 loss,~$\mathcal{L}_{LPIPS}$ is the LPIPS loss~\cite{LPIPS} around the masked region,~$I_{In}$ is the inpainted 2D image and~$I$ is the current~3D scene projected into the respective view~$v$ in 2D.

The depth loss is described by
\begin{equation}
    \mathcal{L}_{depth} = \mathcal{L}^{M}_{L1}(I^{D}_{In}, I^{D}),
\end{equation}
with~$I^{D}_{In}$ and $I^{D}$ as the inpainted depth image and the current projected depth image of the 3D scene, respectively.

The total loss equates to 
\begin{equation}
    \mathcal{L} = \mathcal{L}_{3D} + \mathcal{L}_{id} + \mathcal{L}_{recon} + \mathcal{L}_{depth},
\end{equation}
where~$\mathcal{L}_{3D}$ and~${\mathcal{L}_{id}}$ are two losses from Gaussian Grouping~\cite{gaussian_grouping}.
Since the background already has a high quality before finetuning, we freeze the weights of the Gaussians from the original scene and only optimize over the newly added ones in the inpainted region.

\section{Experiments and Ablations}
We provide additional material for the 2D inpainers and the qualitative and quantitative results.

\subsection{2D Inpainter Comparison}
\textbf{LaMa}. Many 3D inpainting pipelines~\cite{3dgic, gaussian_grouping, inpaint360gs} use LaMa~\cite{LaMa} as their 2D inpainter.
It produces smooth results without sharp details in the inpainted region, which is beneficial in the 3D setting, since it is easier to guarantee multi-view consistency with these smooth inpaintings.
While modern inpainters~\cite{powerpaint, brushnet} often require a detailed and tailored text prompt as guidance, LaMa~\cite{LaMa} only uses an inpainting mask and an image.
Empirically, we found that LaMa~\cite{LaMa} requires masks significantly larger than the hole to achieve high quality inpainting results. Because of this, we increase the size of the masks by dilating the region.
Importantly, we use the same mask for LaMa~\cite{LaMa}, PowerPaint~\cite{powerpaint}, and BrushNet~\cite{brushnet} to keep the results comparable.

\textbf{PowerPaint}. A more recent diffusion-based approach is PowerPaint~\cite{powerpaint}. We specifically use the object removal version of the model. It uses a positive and negative text prompt for guidance as well as an inpainting mask. Our text prompts and hyperparameter details can be found in \cref{subsec:pp_brushnet_prompt}.
While the diffusion-based model~\cite{powerpaint} creates higher fidelity inpaintings than LaMa~\cite{LaMa}, it increases the risk of unwanted hallucinations and degrades the background quality.
To overcome the latter limitation, we replace the unmasked region of the respective image with the input image after inpainting. Since the input image and the inpainted version have slightly different color tones, the masked region is often visible after inpainting.

\textbf{Nano Banana}. The text-to-image model, Nano Banana~\cite{nanobanana}, produces inpainted images of high quality. We only use Nano Banana~\cite{nanobanana} to inpaint the to-be-removed object directly as described in~\cref{subsec:method_scratch}.
As the object is in the input image and we do not use an inpainting mask, we write a scene-specific text prompt. The exact text prompt for every scene is in \cref{subsec:nano_prompt_details}.

\textbf{BrushNet}. The diffusion-based model, BrushNet~\cite{brushnet}, takes a mask and a text prompt as input for inpainting. As it is primarily trained on creating new foreground objects, it struggles to create plain background inpaintings and is therefore not well-suited for object removal. 
Due to these major limitations within the object removal domain, we do not include this model in our 3D evaluation of the scene.

\textbf{Prompt Impact}. We input the same positive and negative input prompts to BrushNet~\cite{brushnet} and PowerPaint~\cite{powerpaint}. However, BrushNet~\cite{brushnet} adds a new object to the scene instead of inpainting the masked area with a similar style as the space around it. Since the word "mask" appeared in the text prompt, the inpainted region shows a mask in the kitchen and living room scene.

\textbf{Depth Comparison}. In \cref{fig:depth_inpainter_comparison}, the depth output of the LaMa~\cite{LaMa} and the PowerPaint~\cite{powerpaint} model are depicted. The LaMa model~\cite{LaMa} excels at this inpainting task, since the task requires a smooth inpainting that matches the depth around the mask. PowerPaint~\cite{powerpaint} struggles with the depth inpainting as the tone difference between the input image and the inpainted region is large. This can be seen as the masked region is clearly darker than the surrounding area.

\input{full_figures/depth_inpainter_comparison}

\subsection{Quantitative Evaluation}
To calculate the metrics, we use the ground truth inpainting mask and ground truth inpainted result from the respective dataset. For the living room scene, we use the inpainting mask calculated by 3DGIC~\cite{3dgic} and the closest viewpoint image from the ground truth.
Across all three scenes, we allocated approximately 90\% of the views for training and 10\% for testing. For every method, we take the average score of all three scenes. We also provide a separate table with the metrics for every scene.

\textbf{Number of Gaussians}. When comparing the number of Gaussians in a scene, the approaches can be divided into two groups. The methods that initialize the scene from scratch output around~$860.000$ Gaussians. The four finetuning methods create scenes with approximately~$940.000$ Gaussians. The 2D inpainter has no noticeable effect on the number of Gaussians created.

\textbf{Run Time}. Conversely, the runtime of the methods depends on the 2D inpainter. While PowerPaint~\cite{powerpaint} takes on average~37 minutes to inpaint every image in a scene, LaMa~\cite{LaMa} takes on average~71 minutes. Overall, the finetuning approaches are faster than \InitLaMa and \InitPowerPaint. This is because finetuning takes less time than running Gaussian Grouping~\cite{gaussian_grouping} from scratch again.
Since \InitNanoBanana only runs Gaussian Grouping~\cite{gaussian_grouping} once without object removal or finetuning, the method is the fastest with just under~1.5 hours. Importantly, the 2D inpainting time of Nano Banana~\cite{nanobanana} is not included in this measurement, since we cannot record the run time of the external API of Nano Banana~\cite{nanobanana}.
These runtime tests are conducted on an NVIDIA RTX 6000 Ada Generation with~48GB of VRAM and~18.176 cores.

\subsection{3D Inpainter Comparison}

\textbf{Background Artifacts}. The background artifacts in the four finetuning methods are not placed at the same depth as in the original scene but much closer to the camera position. This limitation can be explained by considering the depth images with the hole. An example from the kitchen scene is given in \cref{fig:min_max_depth}.
The global depth range results in a higher global precision, including the background depth, but fails to show the depth of the to-be-inpainted hole. Because of this limitation,~3DGIC~\cite{3dgic} and our finetuning methods inpaint the depth images with the manual depth range. While this manual depth representation displays the depth of the hole in the foreground more accurately and therefore gives higher fidelity inpainting results, the background features all lie in one plane. As a consequence, our finetuning methods fail to place the background Gaussians in the correct depth.
\input{full_figures/min_max_depth}

\textbf{Bear COLMAP Ablation}. Since we do not use COLMAP~\cite{colmap} again before running Gaussian Grouping~\cite{gaussian_grouping} in \Init and \InitNanoBanana, the depth of the original bear scene is used. Of the three methods, \InitNanoBanana produces the most artifacts within the depth of the bear, which can be seen in the top right of \cref{fig:3D_bear}, where the outline of the bear is still visible because of the green artifacts. The blurry results in the bottom row of the figure demonstrate why we do not run COLMAP~\cite{colmap} on the inpainted images.
We assume the blur is because the main object is removed and COLMAP~\cite{colmap} has fewer clear features to align the input images with.

\input{full_figures/3D_bear}

\textbf{Living Room PowerPaint Artifacts}. The 2D artifacts of PowerPaint~\cite{powerpaint} propagate to 3D, which can be seen in \cref{fig:powerpaint_artifacts}. The yellow artifact of the first view is visible from nearby views in 3D. Since PowerPaint~\cite{powerpaint} generates many artifacts, the inpainted region produces lower quality results than comparable outputs of methods using LaMa~\cite{LaMa}.

\input{full_figures/powerpaint_artifacts}

\textbf{Quantitative and Qualitative Comparison}. The quantitative results from \cref{subsec:quan_eval} are reflected in the visual qualitative results. Firstly, the models using LaMa~\cite{LaMa} as an inpainter produce higher quality results than those using PowerPaint~\cite{powerpaint} since the latter hallucinate many artifacts.

The quantitative difference between \FTLaMaThree and \FTLaMaAll is underlined by the visual results, especially when considering the kitchen scene. While both approaches struggle with the hole on the blue place mat, \FTLaMaAll produces a significantly higher inpainting quality. The benefit of using all views as reference views is having an RGB loss over every viewpoint, leading to a higher fidelity in training views. However, this approach does not guarantee cross-view consistency.

Another observation from \cref{subsec:quan_eval} is that the \Init methods produce better results than the finetuning methods. In all three scenes, both the background preservation quality and the inpainting quality are higher in the \Init results. When using LaMa~\cite{LaMa}, this difference is not as significant. Especially \FTLaMaAll produces results very similar to \InitLaMa.

It is important to note that our results do not display the same 3D inpainting quality as the~3DGIC paper~\cite{3dgic}. We assume this is because we focus on single-step approaches by only inpainting each view once and because we do not employ the same cross-view consistent loss (see Eq. 9 of~\cite{3dgic}).

\section{Prompt Details}
\label{sec:prompt_details}
For reproduction, we specify the exact prompts we use for the respective experiments.

\subsection{Nano Banana Prompt}
\label{subsec:nano_prompt_details}
When inpainting with Nano Banana~\cite{nanobanana}, we use a scene specific prompt because each scene has a different object to remove. They are given in the following. \\
\textbf{Bear}. Remove the grey stone bear sculpture and the square grey stone plinth beneath it from the scene. This includes the entire bear figure and its base. Inpaint the area by completing and extending the texture of the large, irregular stone slab, ensuring it appears as a single, continuous, and naturally textured surface. Extend the surrounding background environment, specifically the large redwood tree trunks, the green ivy plants, the surrounding leaf litter, and dirt, seamlessly into the space previously occupied by the statue and its base, ensuring they are not distorted. Ensure the overall natural woodland atmosphere and perspective are preserved. Remove all contact shadows, complex shadows, and any color-bleed reflections cast by the removed objects on the stone slab and the forest floor. The small white object on the stone slab must remain in its position. \\
\textbf{Kitchen}. Remove the yellow Technic bulldozer toy from the table. This includes the main body, the black treads, the mechanical arms, and the front bucket. Inpaint the area by extending the textures of the grey woven placemat and the light pink fabric mat underneath it. Ensure the wood grain of the table visible in the background is reconstructed seamlessly where the toy was. Remove all contact shadows and yellow color-bleed reflections on the mats and table surface. \\
\textbf{Living room}. Remove the white fluffy plush creature with horns from the table. This includes its fuzzy white body, brown horns, pink tongue, and facial features. Inpaint the substantial area it occupied by seamlessly extending the light-wood grain texture of the table surface. Critically, preserve all surrounding objects in their precise locations. Remove all contact shadows and any light reflections cast by the creature. The new, continuous wood surface should have consistent, natural contact shadows for the remaining objects.

\subsection{PowerPaint and BrushNet Prompt}
\label{subsec:pp_brushnet_prompt}
For the PowerPaint~\cite{powerpaint} and BrushNet~\cite{brushnet} inpainting results, we use a generic prompt for inpainting the hole in the scene. Additionally, we have a negative prompt to avoid artifacts. Both of them are used for all three scenes. \\
\textbf{Positive prompt}. Restore only the masked region to match the surrounding scene. Keep lighting, colors, texture, geometry, and perspective consistent. Do not alter any unmasked pixels. Seamless boundary blend. \\
\textbf{Negative prompt}. changes outside mask, global color shift, exposure shift, contrast shift, style change, camera change, composition change, warped edges, blur, artifacts.

PowerPaint~\cite{powerpaint} has additional hyperparameters. For the number of steps, we use~45. We set the guidance scale to~20 for inpainting RGB images and~10 for inpainting depth images.

\input{full_figures/table_bear}
\input{full_figures/table_kitchen}
\input{full_figures/table_living_room}

%% file: full_figures/depth_inpainter_comparison.tex
\begin{figure}[t]
    \centering
    \begin{subfigure}[b]{0.155\textwidth}
         \centering
         \includegraphics[width=\textwidth]{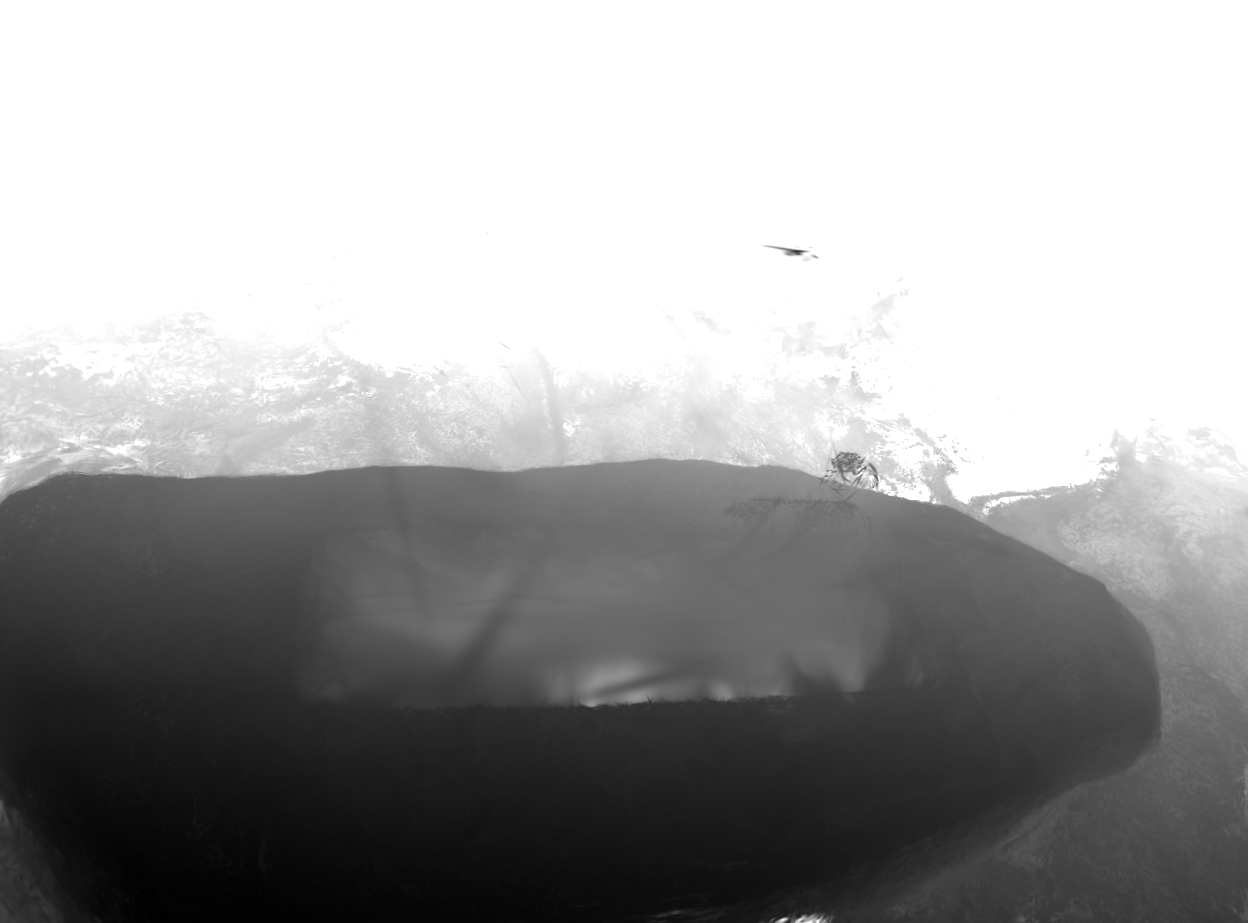}
         \caption{Input}
     \end{subfigure}
      \hfill
    \begin{subfigure}[b]{0.155\textwidth}
        \centering
        \includegraphics[width=\textwidth]{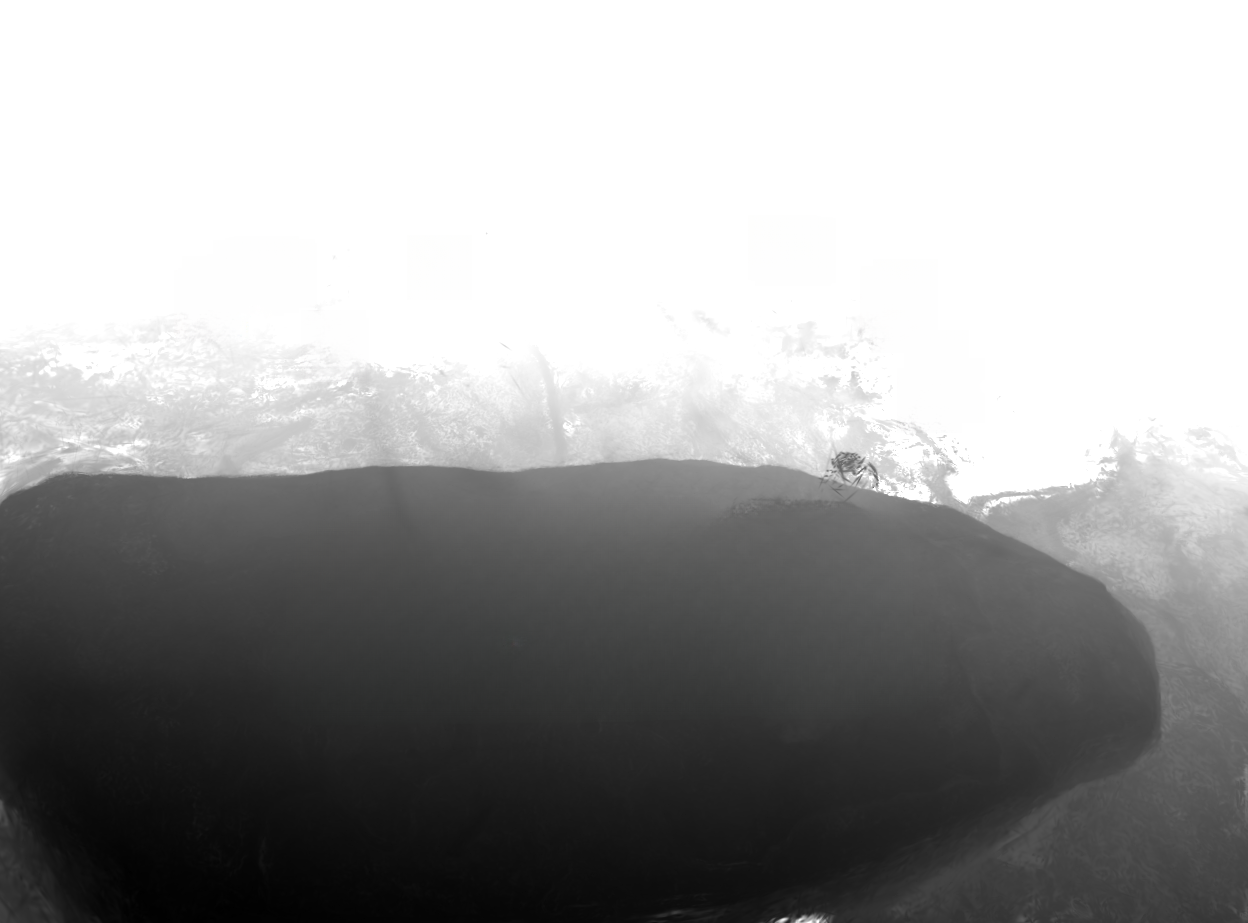}
        \caption{LaMa~\cite{LaMa}}
    \end{subfigure}
    \hfill
     \begin{subfigure}[b]{0.155\textwidth}
         \centering
         \includegraphics[width=\textwidth]{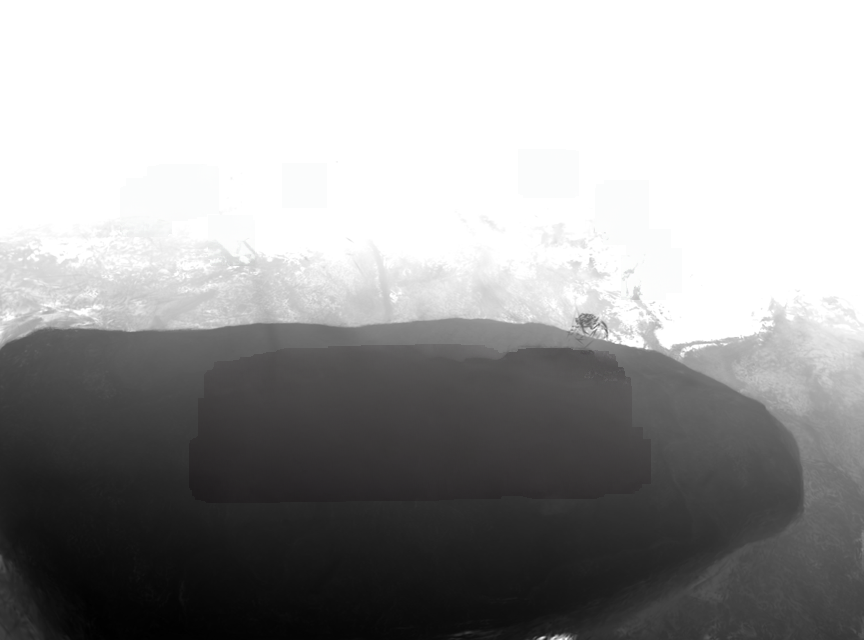}
         \caption{PowerPaint~\cite{powerpaint}}
     \end{subfigure}

     \caption{\textbf{2D Depth Inpainting Comparison}. On the left is the input of the depth image after object removal. We compare the LaMa~\cite{LaMa} inpainting results with the PowerPaint~\cite{powerpaint} output.}
     \label{fig:depth_inpainter_comparison}
\end{figure}

%% file: full_figures/min_max_depth.tex
\begin{figure}[t]
    \centering
    \begin{subfigure}[b]{0.23\textwidth}
         \centering
         \includegraphics[width=\textwidth]{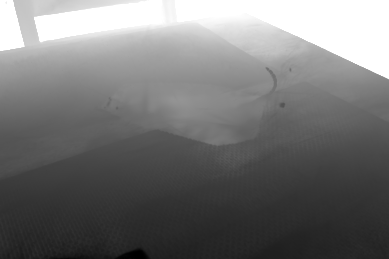}
         \caption{Manual foreground depth range}
     \end{subfigure}
      \hfill
    \begin{subfigure}[b]{0.23\textwidth}
        \centering
        \includegraphics[width=\textwidth]{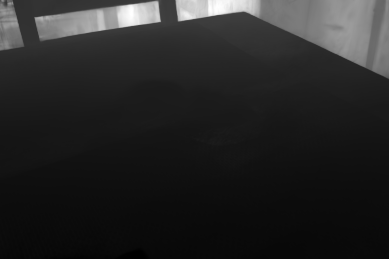}
        \caption{Global depth range}
    \end{subfigure}
     \caption{\textbf{Depth Representation Comparison in the Kitchen Scene}. While 3DGIC~\cite{3dgic} uses the left depth range for a higher fidelity in inpainting the hole, the right depth range produces a higher background quality with its global depth representation.}
     \label{fig:min_max_depth}
\end{figure}

%% file: full_figures/3D_bear.tex
\begin{figure}[t]
     \centering
     \begin{subfigure}[b]{0.155\textwidth}
         \centering
         \includegraphics[width=\textwidth]{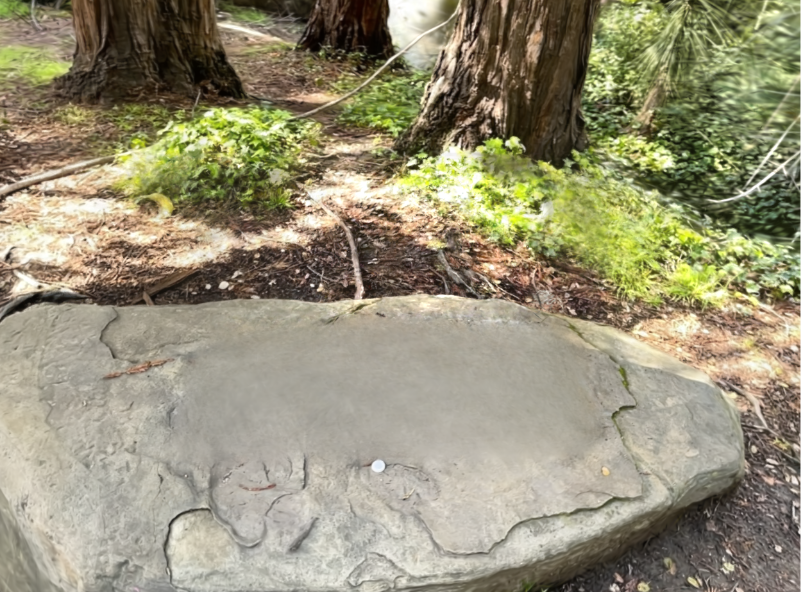}
     \end{subfigure}
      \hfill
     \begin{subfigure}[b]{0.155\textwidth}
         \centering
         \includegraphics[width=\textwidth]{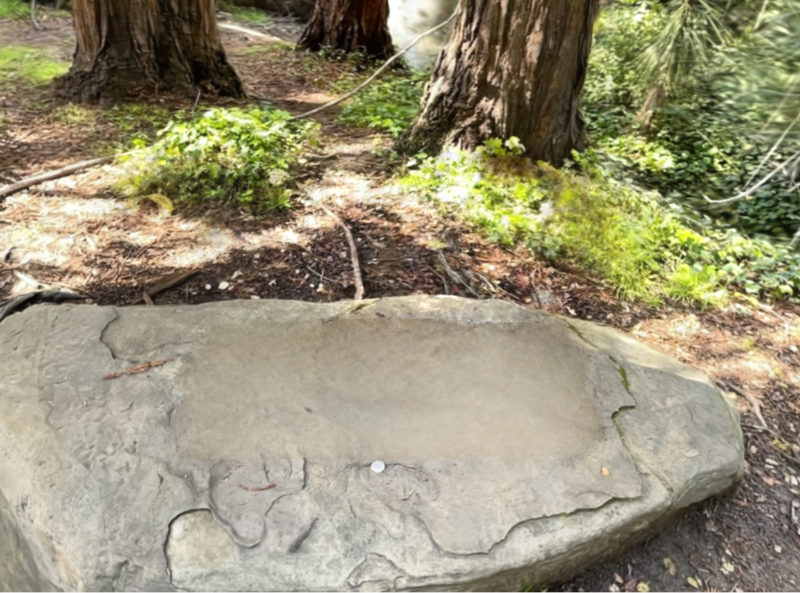}
     \end{subfigure}
     \hfill
     \begin{subfigure}[b]{0.155\textwidth}
         \centering
         \includegraphics[width=\textwidth]{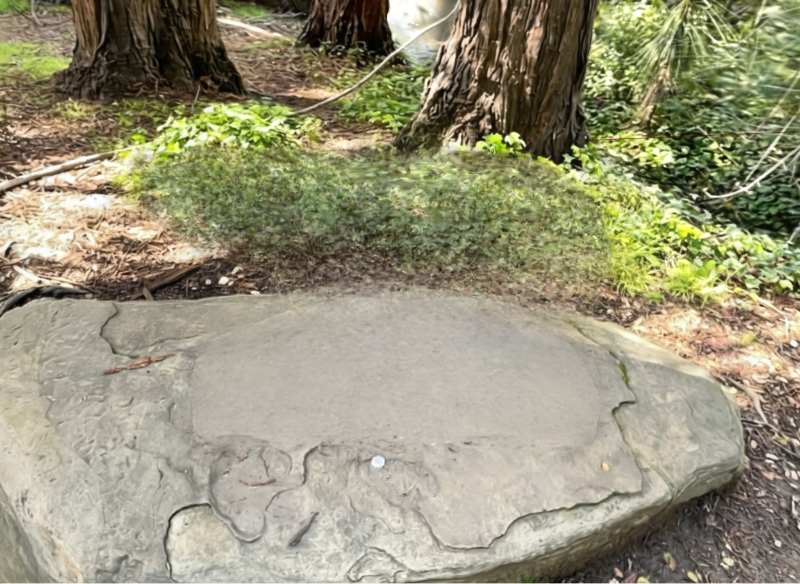}
     \end{subfigure}

     \vspace{0.1cm} 

     \begin{subfigure}[b]{0.155\textwidth}
         \centering
         \includegraphics[width=\textwidth]{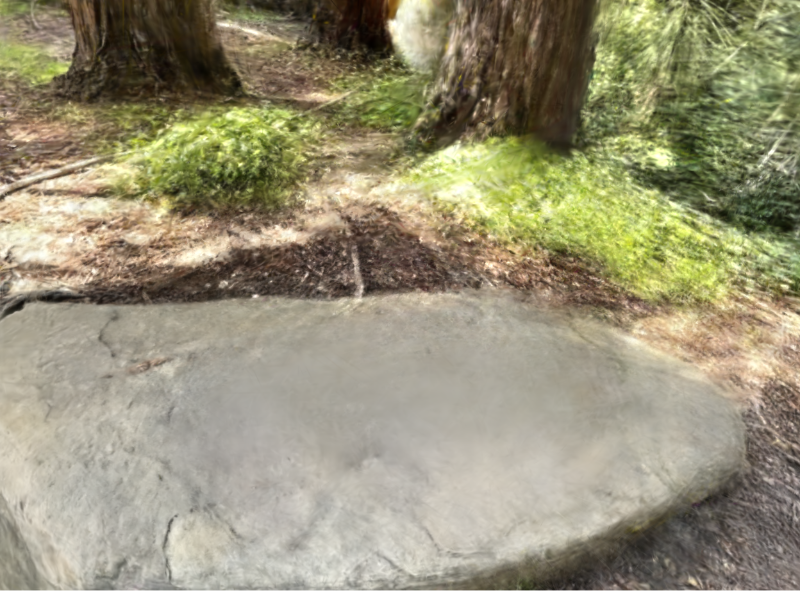}
         \caption{\InitLaMaShort}
     \end{subfigure}
      \hfill
     \begin{subfigure}[b]{0.155\textwidth}
         \centering
         \includegraphics[width=\textwidth]{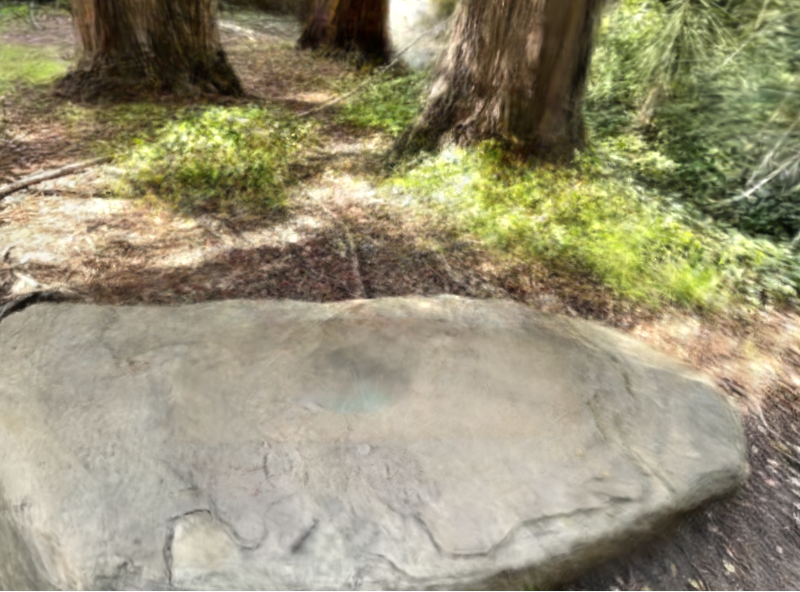}
         \caption{\InitPowerPaintShort}
     \end{subfigure}
     \hfill
     \begin{subfigure}[b]{0.155\textwidth}
         \centering
         \includegraphics[width=\textwidth]{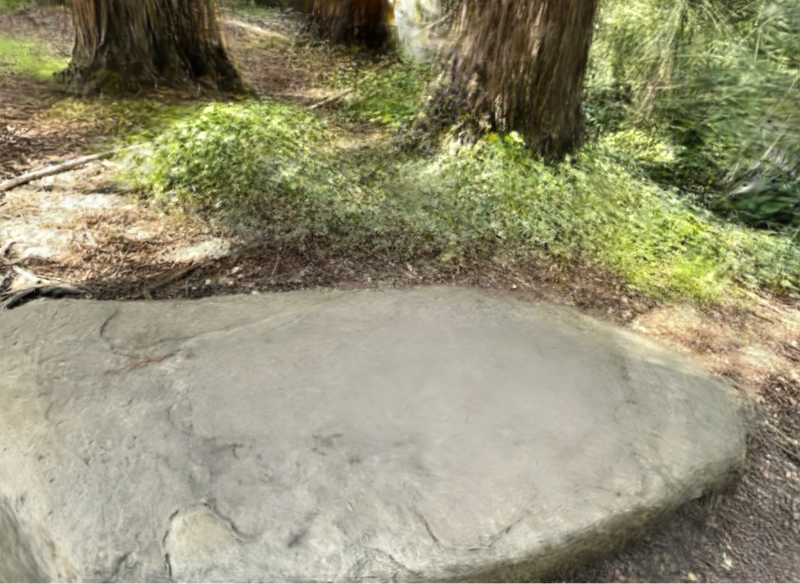}
         \caption{\InitNanoBananaShort}
     \end{subfigure}

     \caption{\textbf{Ablation Study.} On the top, the respective method is run without running COLMAP~\cite{colmap} again giving artifacts in the depth of the original bear. On the bottom, the respective method is run with an additional COLMAP~\cite{colmap} call giving blurry results.}
     \label{fig:3D_bear}
\end{figure}

%% file: full_figures/powerpaint_artifacts.tex
\begin{figure}[t]
     \centering
     \begin{subfigure}[b]{0.235\textwidth}
         \centering
         \includegraphics[width=\textwidth]{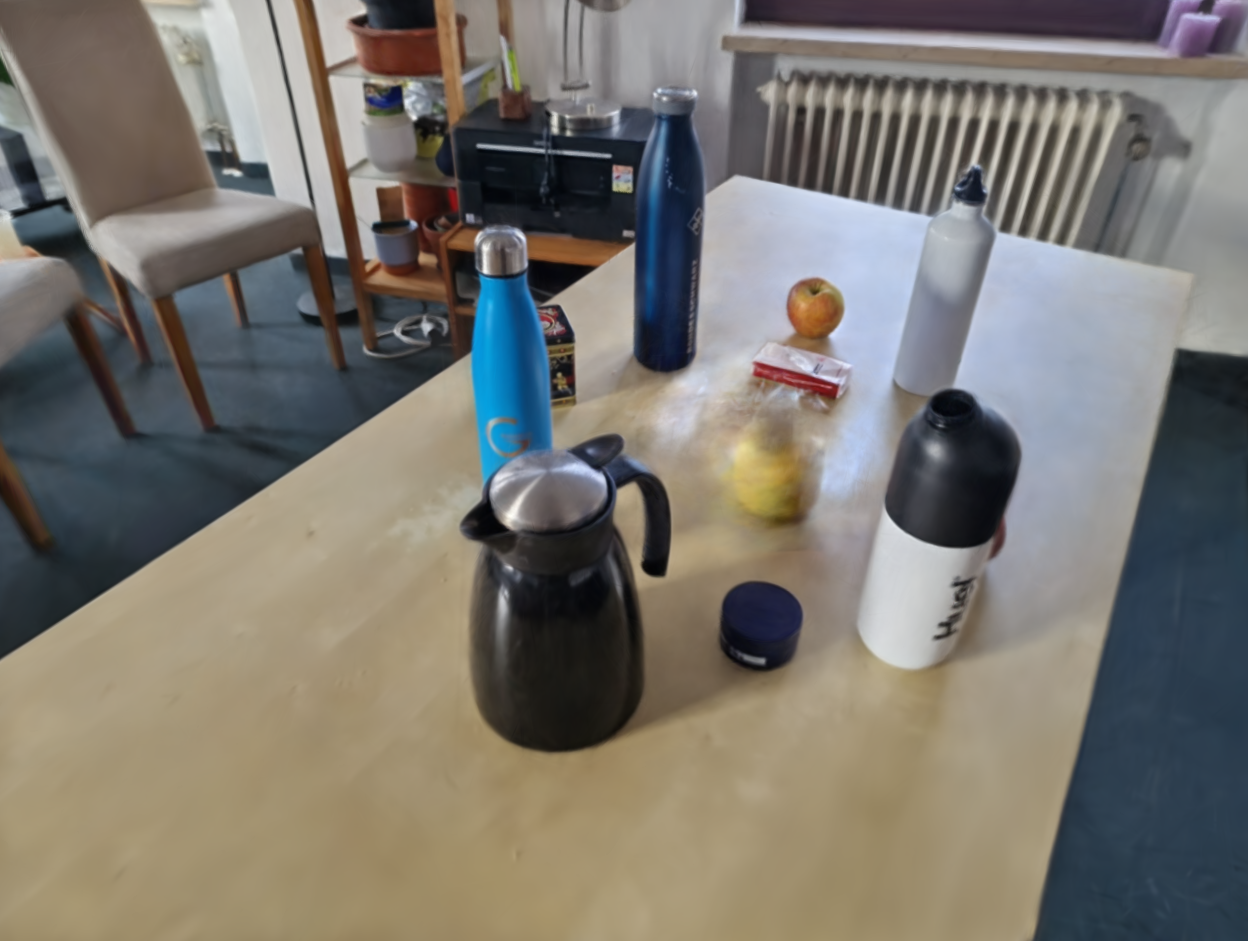}
     \end{subfigure}
      \hfill
     \begin{subfigure}[b]{0.235\textwidth}
         \centering
         \includegraphics[width=\textwidth]{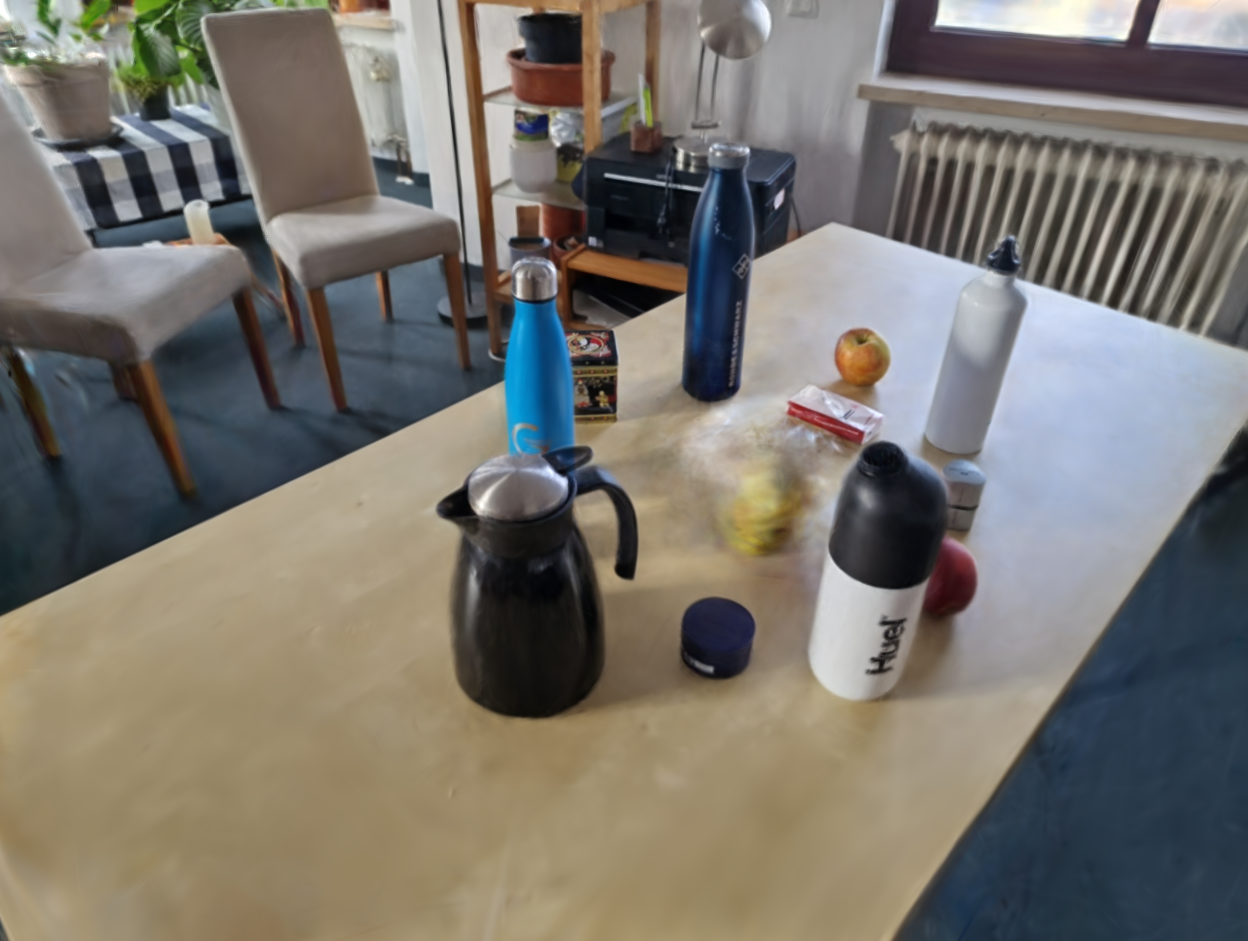}
     \end{subfigure}

     \vspace{0.1cm} 

     \begin{subfigure}[b]{0.235\textwidth}
         \centering
         \includegraphics[width=\textwidth]{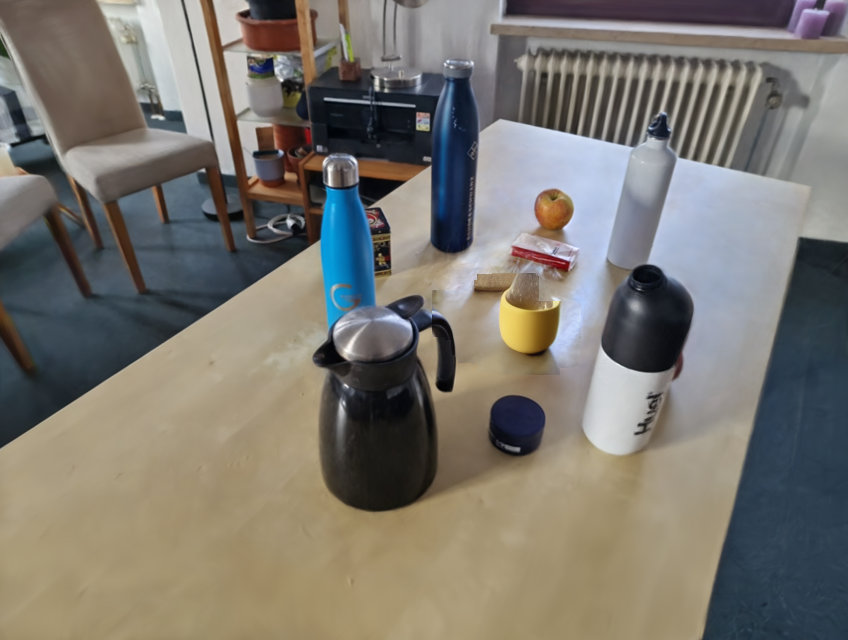}
         \caption{View A}
     \end{subfigure}
      \hfill
     \begin{subfigure}[b]{0.235\textwidth}
         \centering
         \includegraphics[width=\textwidth]{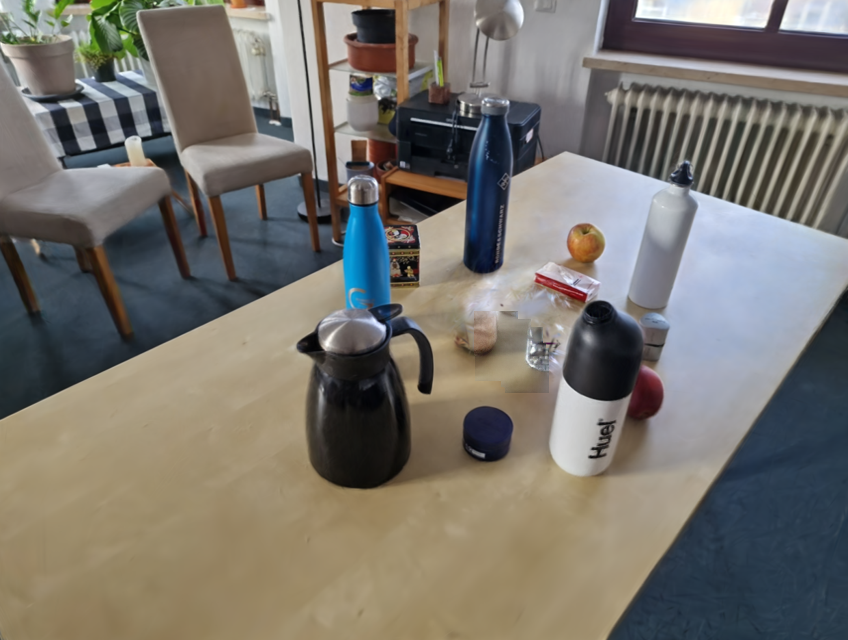}
         \caption{View B}
     \end{subfigure}

     \caption{\textbf{Artifacts in 3D with PowerPaint}. The 3D results of \InitPowerPaint are presented (top) with the respective 2D inpainting results of PowerPaint~\cite{powerpaint} (bottom). The yellow artifact in view A is visible in the 3D results of both view A and view B.}
     \label{fig:powerpaint_artifacts}
\end{figure}

%% file: full_figures/table_bear.tex
\begin{table*}[t]
    \centering
    \begin{tabular}{@{}llllllllll@{}}
        \toprule
        Method & PSNR $\uparrow$ & m-PSNR $\uparrow$ & SSIM $\uparrow$ & m-SSIM $\uparrow$ & LPIPS $\downarrow$ & m-LPIPS $\downarrow$ & FID $\downarrow$ & m-FID $\downarrow$ & Run time \\ 
        \midrule
        \FTLaMaThreeShort & 16.45 & 20.43 & 0.6157 & 0.9201 & 0.411 & 0.101 & 157.91 & \textbf{160.78} & 3h 31min \\
        \FTPowerPaintThreeShort & 14.93 & 19.32 & 0.5613 & 0.9243 & 0.4762 & 0.1098 & 327.81 & 303.66 & 3h 4min \\
        \FTLaMaAllShort & 18.03 & 22.21 & 0.6509 & \textbf{0.9404} & 0.3982 & 0.0939 & 210.21 & 210.47 & 3h 30min \\
        \FTPowerPaintAllShort & 13.77 & 20.64 & 0.4192 & 0.9181 & 0.5771 & 0.1105 & 314.19 & 177.12 & 3h 7min \\
        \InitLaMaShort & \textbf{19.65} & \textbf{23.5} & \textbf{0.7026} & 0.9327 & 0.3363 & \textbf{0.0878} & \textbf{138.62} & 179.96 & 4h 26min \\
        \InitPowerPaintShort & \textbf{19.65} & 21.86 & 0.6991 & 0.9243 & \textbf{0.3268} & 0.0893 & 155.08 & 222.9 & 4h 5min \\
        \InitNanoBananaShort & 14.9 & 19.61 & 0.3737 & 0.9143 & 0.4583 & 0.0942 & 198.44 & 207.23 & \textbf{1h 26min} \\
        \bottomrule
    \end{tabular}
    \caption{\textbf{Quantitative Evaluation of the 3D Inpainting Results on the Bear Scene}. The masked metrics (e.g., m-PSNR) represent the scores calculated only within the ground truth inpainting mask. The best score is highlighted in bold font. Note that the run time of \InitNanoBananaShort\xspace does not include the 2D inpainting run time of the external API.}
    \label{tab:bear_eval}
\end{table*}

%% file: full_figures/table_kitchen.tex
\begin{table*}[t]
    \centering
    \begin{tabular}{@{}llllllllll@{}}
        \toprule
        Method & PSNR $\uparrow$ & m-PSNR $\uparrow$ & SSIM $\uparrow$ & m-SSIM $\uparrow$ & LPIPS $\downarrow$ & m-LPIPS $\downarrow$ & FID $\downarrow$ & m-FID $\downarrow$ & Run time \\ 
        \midrule
        \FTLaMaThreeShort & 24.63 & 18.67 & 0.836 & 0.9084 & 0.2031 & 0.0718 & 79.47 & 122.19 & 6h \\
        \FTPowerPaintThreeShort & 21.64 & 16.99 & 0.7832 & 0.9019 & 0.2611 & 0.0812 & 205.72 & 172.99 & 4h 46min \\
        \FTLaMaAllShort & 25.85 & 19.92 & 0.8637 & 0.9237 & 0.1824 & 0.0595 & 83.84 & 90.49 & 6h 3min \\
        \FTPowerPaintAllShort & 19.6 & 17.66 & 0.6693 & 0.8966 & 0.3988 & 0.0887 & 278.33 & 168.16 & 4h 50min \\
        \InitLaMaShort & 26.42 & 19.94 & \textbf{0.8713} & 0.923 & \textbf{0.1662} & \textbf{0.0559} & 86.57 & 85.26 & 7h 2min \\
        \InitPowerPaintShort & \textbf{27.01} & 20.87 & 0.8626 & \textbf{0.9259} & 0.1867 & 0.0633 & 91.14 & 91.51 & 5h 51min \\
        \InitNanoBananaShort & 26.37 & \textbf{22.58} & 0.7709 & 0.9199 & 0.2328 & 0.0561 & \textbf{59.81} & \textbf{69.4} & \textbf{1h 26min} \\
        \bottomrule
    \end{tabular}
    \caption{\textbf{Quantitative Evaluation of the 3D Inpainting Results on the Kitchen Scene}. The masked metrics (e.g., m-PSNR) represent the scores calculated only within the ground truth inpainting mask. The best score is highlighted in bold font. Note that the run time of \InitNanoBananaShort\xspace does not include the 2D inpainting run time of the external API.}
    \label{tab:kitchen_eval}
\end{table*}

%% file: full_figures/table_living_room.tex
\begin{table*}[t]
    \centering
    \begin{tabular}{@{}llllllllll@{}}
        \toprule
        Method & PSNR $\uparrow$ & m-PSNR $\uparrow$ & SSIM $\uparrow$ & m-SSIM $\uparrow$ & LPIPS $\downarrow$ & m-LPIPS $\downarrow$ & FID $\downarrow$ & m-FID $\downarrow$ & Run time \\ 
        \midrule
        \FTLaMaThreeShort & 12.43 & 16.13 & 0.6123 & 0.9846 & 0.6267 & 0.0261 & 140.99 & 264.26 & 3h 34min \\
        \FTPowerPaintThreeShort & \textbf{12.73} & 17.2 & \textbf{0.6496} & 0.9889 & 0.6398 & 0.025 & 192.01 & 212.89 & 3h 9min \\
        \FTLaMaAllShort & 12.39 & 16.63 & 0.6163 & 0.9905 & \textbf{0.6242} & 0.023 & 134.58 & 232.05 & 3h 39min \\
        \FTPowerPaintAllShort & 12.67 & 15.74 & 0.5841 & 0.9809 & 0.6418 & 0.0291 & 282.79 & 292.83 & 3h 10min \\
        \InitLaMaShort & 12.16 & 14.49 & 0.6163 & 0.9894 & 0.6276 & 0.0231 & 128.46 & 196.19 & 4h 40min \\
        \InitPowerPaintShort & 12.38 & 15.59 & 0.6218 & 0.9901 & 0.6307 & 0.0241 & \textbf{124.48} & 221.46 & 4h 24min \\
        \InitNanoBananaShort & 12.37 & \textbf{17.21} & 0.6149 & \textbf{0.9917} & 0.6311 & \textbf{0.0222} & 137.34 & \textbf{184.84} & \textbf{1h 28min} \\
        \bottomrule
    \end{tabular}
    \caption{\textbf{Quantitative Evaluation of the 3D Inpainting Results on the Living Room Scene}. The masked metrics (e.g., m-PSNR) represent the scores calculated only within the ground truth inpainting mask. The best score is highlighted in bold font. Note that the run time of \InitNanoBananaShort\xspace does not include the 2D inpainting run time of the external API.}
    \label{tab:living_room_eval}
\end{table*}